\documentclass{article}

\usepackage[main, final]{NIPS2026-CameraReady/neurips_2026}

\usepackage[utf8]{inputenc} 
\usepackage[T1]{fontenc}    
\usepackage{hyperref}       
\usepackage{url}            
\usepackage{booktabs}       
\usepackage{amsfonts}       
\usepackage{nicefrac}       
\usepackage{microtype}      
\usepackage{xcolor}         

\usepackage{graphicx}
\usepackage{subcaption}
\usepackage{amsmath}
\usepackage{amssymb}
\usepackage{mathtools}
\usepackage{amsthm}
\usepackage[capitalize,noabbrev]{cleveref}

\theoremstyle{plain}

\theoremstyle{definition}

\theoremstyle{remark}

\usepackage[textsize=tiny]{todonotes}

\newif\ifshowcomment
\showcommentfalse 
\usepackage{wrapfig}
\usepackage{multirow}
\usepackage{makecell}
\usepackage{xspace}
\usepackage{enumerate}
\usepackage{listings}
\usepackage{algorithmic}
\usepackage{algorithm}
\usepackage{mathrsfs}
\usepackage{natbib}
\setcitestyle{numbers,square}

\usepackage{paralist, tabularx}
\usepackage{inconsolata}
\usepackage{epigraph}
\usepackage{float}
\usepackage[normalem]{ulem}
\useunder{\uline}{\ul}{}
\usepackage[framemethod=TikZ]{mdframed}

\usepackage{framed}
\usepackage{array}
\usepackage{verbatim} 
\usepackage{bbm}
\usepackage{commath}
\usepackage{amsbsy}
\usepackage[shortlabels]{enumitem}
\usepackage{tcolorbox}
\usepackage{bbding}
\usepackage{threeparttable}
\usepackage{lipsum}
\usepackage{minitoc}
\usepackage{tocloft}
\usepackage[export]{adjustbox}

\newcommand{\model}{ReST-RL\xspace}
\newcommand{\modelTrain}{ReST-GRPO\xspace}
\newcommand{\modelTest}{VM-MCTS\xspace}
\newcommand{\modelGRPO}{GRPO\xspace}
\newcommand{\modelDPO}{ReST-DPO\xspace}
\newcommand{\modelORM}{ORM-MCTS\xspace}
\newcommand{\modelSFT}{ReST$^\text{EM}$\xspace}
\newcommand{\vpara}[1]{\noindent\textbf{#1}\xspace}
\newcommand{\hhide}[1]{}

\newcommand{\pycomment}[1]{\textcolor{gray!70}{\# #1}}

\title{ReST-RL: Reinforcing LLM Reasoning through Unified Self-Training and Value-Guided Search}

\author{%
  Sining Zhoubian$^{\ast}$, Dan Zhang$^{\dagger}$, Jie Tang$^{\dagger}$ \\ 
  \textmd{Department of Computer Science and Technology, Tsinghua University} \\
  \texttt{zbsn21@mails.tsinghua.edu.cn} \\
  \url{https://github.com/THUDM/ReST-RL}
}

\begin{document}

\maketitle

\renewcommand{\thefootnote}{\fnsymbol{footnote}}
\footnotetext[1]{Work done when interned at Z.ai.}
\footnotetext[2]{Corresponding: DZ and JT}

\begin{abstract}
With respect to improving the reasoning accuracy of LLMs, the representative reinforcement learning (RL) method --- Group Relative Policy Optimization (GRPO) --- has achieved critical success, yet it still suffers from the issue of insignificant reward signals.
This paper introduces \textbf{\model}, a unified Reinforced Self-Training (ReST) policy--value framework that reconnects policy optimization and value-guided search to improve LLM reasoning ability.
Firstly, \textit{\modelTrain} adopts an optimized ReST-style algorithm to reshape the policy-induced trajectory distribution by increasing the reward variance of GRPO sampling and exposing the policy to more informative partial states, thereby improving training efficiency and effectiveness.
Then, we further introduce a decoding optimization method, \textit{\modelTest}, which trains a Value Model (VM) from self-collected Monte-Carlo Tree Search (MCTS) targets and deploys it through an adapted MCTS algorithm to provide precise process signals and verification scores, further enhancing reasoning accuracy.
These two stages are internally dependent --- \modelTrain yields higher-quality trajectories for value learning with \modelTest, which in turn enables more effective inference-time search.
We validate our framework on multiple coding benchmarks (e.g., APPS, BigCodeBench, and HumanEval), where it significantly outperforms other reinforcement training baselines (naive GRPO, DAPO, and ReST-DPO), as well as decoding and verification baselines (e.g., PRM-BoN and ORM-MCTS), indicating its power to strengthen LLM reasoning capability.
Moreover, we further evaluate \model on out-of-domain math and science reasoning tasks, where it achieves improved performance without target-domain tuning and favorable end-to-end efficiency trade-offs, providing preliminary transfer evidence beyond our primary coding domain.

\end{abstract}

\section{Introduction}
\label{sec: introduction}

Recent advances in Large Language Models (LLMs) have made remarkable progress in reasoning tasks \citep{openai2024gpt4technicalreport,guo2024deepseekcoderlargelanguagemodel,zhang2025lessonsdevelopingprocessreward}, yet substantial challenges remain for solving complex ones \citep{hendrycks2021measuringcodingchallengecompetence,zhuo2025bigcodebenchbenchmarkingcodegeneration}. 
Reinforcement Learning (RL) has emerged as a promising approach to further improve the reasoning ability of LLMs \citep{luo2025wizardmathempoweringmathematicalreasoning,wang2024mathshepherdverifyreinforcellms,schulman2017proximalpolicyoptimizationalgorithms,gulcehre2023reinforcedselftrainingrestlanguage,rafailov2024direct,shao2024deepseekmathpushinglimitsmathematical,singh2023beyond,zhang2024restmctsllmselftrainingprocess}, with growing attention on both online methods like Group Relative Policy Optimization (GRPO) \citep{shao2024deepseekmathpushinglimitsmathematical} that data sampling and model updating take place simultaneously \citep{schulman2017proximalpolicyoptimizationalgorithms,rafailov2024direct,shao2024deepseekmathpushinglimitsmathematical},
and offline self-training paradigms like Reinforced Self-Training (ReST) \citep{gulcehre2023reinforcedselftrainingrestlanguage,zhang2024restmctsllmselftrainingprocess} that obtains training data through offline sampling and filtering mechanisms \citep{gulcehre2023reinforcedselftrainingrestlanguage,singh2023beyond,zhang2024restmctsllmselftrainingprocess}.
In addition, the reward models (RMs) are gradually attracting more attention due to their function in output verification, with outcome reward models (ORMs) that verify the final output of LLMs \citep{uesato2022solvingmathwordproblems,lightman2023letsverifystepstep} and process reward models (PRMs) that provide feedback for intermediate steps and indicate better verification accuracies than ORMs \citep{lightman2023letsverifystepstep,wang2024mathshepherdverifyreinforcellms,zhang2025lessonsdevelopingprocessreward,zhang2024restmctsllmselftrainingprocess}, improving reasoning accuracy of LLMs. 
Moreover, integrating RMs with advanced decoding algorithms, such as Monte-Carlo Tree Search (MCTS), has shown further enhancements in LLM reasoning performance \citep{zhang2024restmctsllmselftrainingprocess,liu2024dontthrowawayvalue}.

\begin{wrapfigure}{r}{0.49\textwidth}
    \vspace{-0.4cm}
    \centering

    \subcaptionbox{
        Reward variance becomes more informative during \modelTrain.
        \label{fig: reward_std}
    }{  \includegraphics[width=0.95\linewidth]{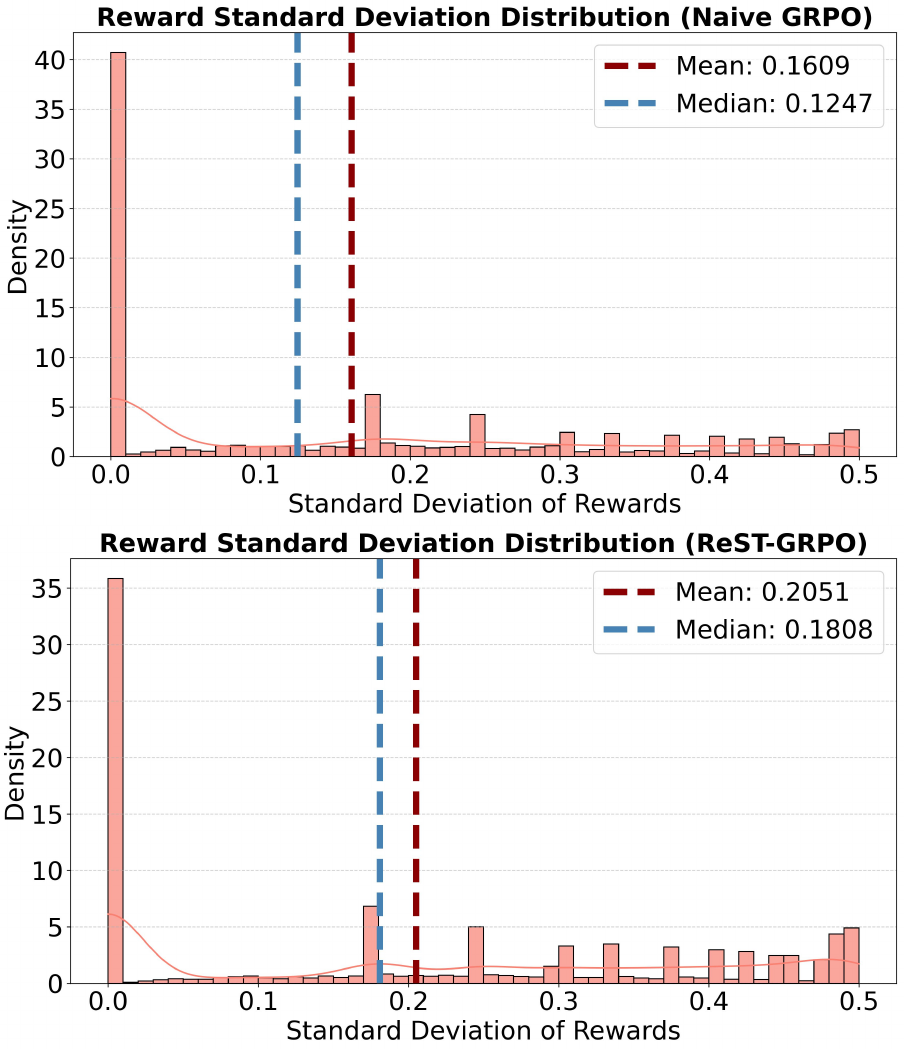}
    }

    \vspace{-0.1cm}

    \subcaptionbox{
        \model improves Qwen3-8B on code and provides transfer evidence on math and science tasks without target-domain tuning.
        \label{fig: intro_performance}
    }{  \includegraphics[width=0.95\linewidth]{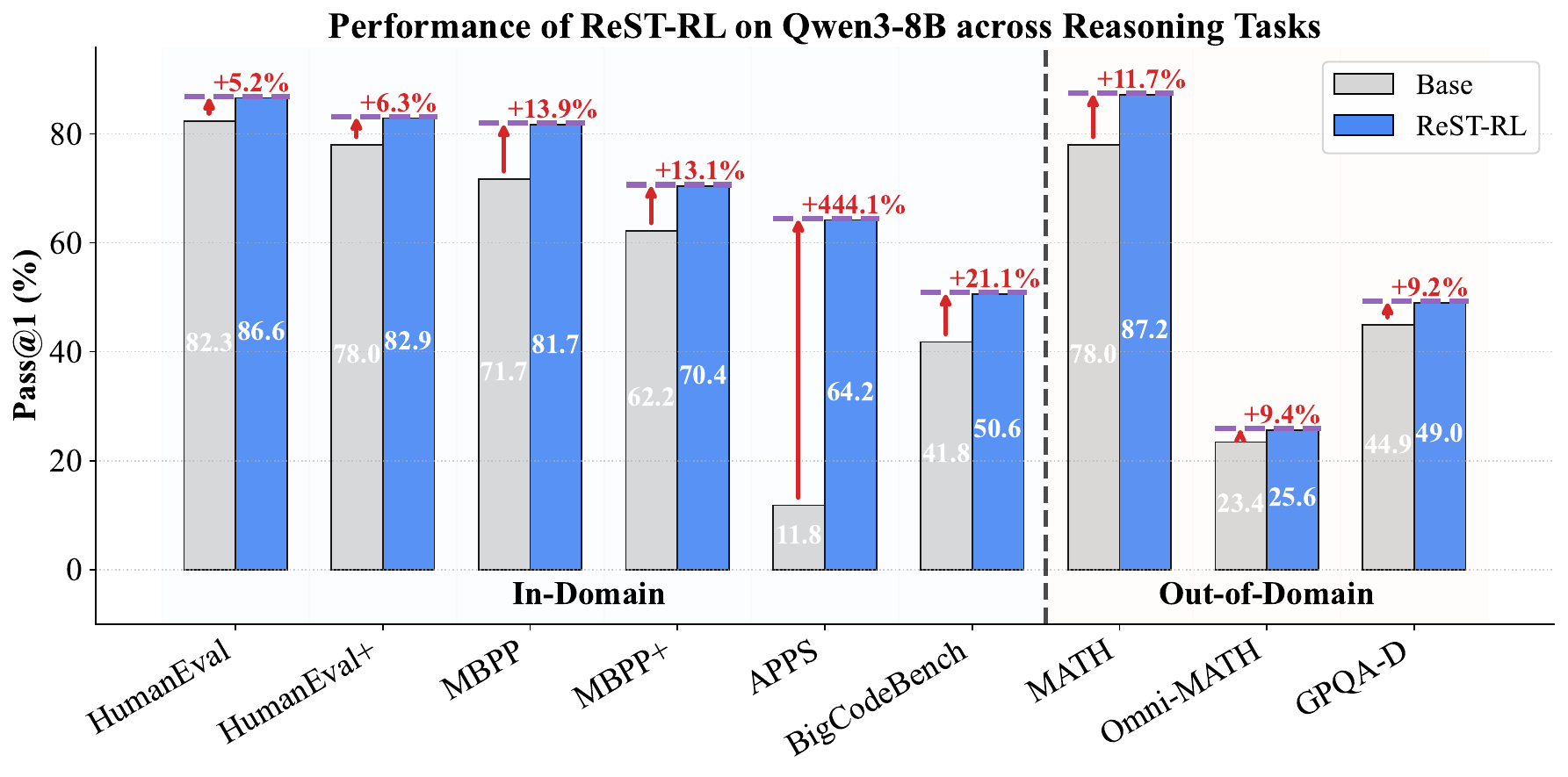}
    }

    \vspace{-0.2cm}
    \caption{Motivation and overall results.}
    \label{fig: intro}
    \vspace{-0.3cm}
\end{wrapfigure}

However, there are still some shortcomings in these methods. 
\textbf{\textit{(1) GRPO can provide low-information updates when all responses in a sampled group receive similar rewards~\citep{yu2025dapoopensourcellmreinforcement,zhang2025r1vllearningreasonmultimodal} (cf. Figure~\ref{fig: reward_std}).}}
Dynamic sampling and step-wise reward designs mitigate this issue \citep{yu2025dapoopensourcellmreinforcement,zhang2025r1vllearningreasonmultimodal}, but may require additional online rollouts or reward design and still operate primarily at the original-prompt level. In particular, DAPO repeatedly samples until enough nonzero-variance groups are retained, whereas a prompt-level curriculum changes which questions are presented. Neither mechanism directly revisits informative intermediate states within a successful trajectory.
\textbf{\textit{(2) A key limitation of PRMs is their reliance on high-quality annotated data~\citep{lightman2023letsverifystepstep,wang2024mathshepherdverifyreinforcellms}, which is costly and labor-intensive to obtain, hindering scalability and reliability.}}
To tackle this problem, some methods propose to estimate and collect process rewards through Monte-Carlo simulations \citep{wang2024mathshepherdverifyreinforcellms,zhang2024restmctsllmselftrainingprocess}.
However, their reliance on task-specific answer matching and fixed notions of step correctness can limit direct application to richer reasoning settings such as coding \citep{zhang2024restmctsllmselftrainingprocess}.
We provide a more comprehensive review of these LLM reinforcement methods (see Appendix \ref{sec: related_work}).
Overall, existing approaches expose a practical trade-off among data-collection cost, informative policy updates, and fine-grained inference-time guidance.
This motivates a design that preserves GRPO's lightweight policy update while recovering informative state-level learning signals. Such a design should distinguish two roles that are often conflated: improving the policy's training distribution, and learning a value signal for allocating inference-time computation. Connecting these roles is useful only if the first-stage policy improvement survives a matched value-learning pipeline rather than merely producing a better policy-only checkpoint.

In this paper, we introduce \model, a unified, state-localized policy--value framework for Reinforced Self-Training, comprising \modelTrain and \modelTest (Figure~\ref{fig: framework}).
Rather than treating policy training and reward learning as two isolated enhancements, the core principle of \model is to reconnect them in a lightweight manner: \modelTrain reshapes the policy-induced trajectory distribution toward more informative high-reward regions, and this improved distribution in turn enables more reliable value estimation and more effective inference-time search in \modelTest.
As the first stage, \modelTrain adopts ReST-style data filtering and assembly to improve \modelGRPO on sparse-reward reasoning tasks. Selected prefixes are not treated as imitation targets: they become additional online-GRPO contexts from which the policy samples and optimizes fresh suffixes. This increases reward variation while retaining exploration over continuations.
As the second stage, \modelTest trains a value model (VM) from self-collected MCTS targets without extra annotation. Under a fixed policy, the VM estimates $V^\pi(s)=\mathbb{E}_\pi[R\mid s]$ and is used both to allocate tree search and to select final outputs.
In this way, \model aims to recover fine-grained value guidance for reasoning without returning to the full cost of online actor-critic learning.
Unlike ReST-style methods that train on accepted complete responses \citep{gulcehre2023reinforcedselftrainingrestlanguage,singh2023beyond}, \modelTrain uses accepted prefixes only as contexts and freshly samples their suffixes online. It also differs from prompt-level curricula and dynamic sampling \citep{parashar2025curriculum,yu2025dapoopensourcellmreinforcement}: these methods change which original prompts are presented, whereas \modelTrain additionally changes where within a trajectory online RL begins.
Our primary evaluation covers coding benchmarks of varying difficulty, including APPS \citep{hendrycks2021measuringcodingchallengecompetence}, BigCodeBench \citep{zhuo2025bigcodebenchbenchmarkingcodegeneration}, and HumanEval \citep{chen2021evaluatinglargelanguagemodels}. We additionally study transfer without target-domain tuning on MATH \citep{hendrycks2021measuringmathematicalproblemsolving}, Omni-MATH \citep{gao2024omnimathuniversalolympiadlevel}, and GPQA-Diamond \citep{rein2023gpqagraduatelevelgoogleproofqa}.
Table~\ref{tab: policy_training} shows improvements over ReST-DPO, GRPO, and DAPO, while Table~\ref{tab: reward_model} compares against ORM-, PRM-, and MCTS-based decoding. The controlled results in Table~\ref{tab: controlled_results} further isolate policy components, matched value learning, and approximately latency-matched search. Together, these results support a coherent policy--value pipeline with favorable end-to-end compute trade-offs and cross-domain transfer evidence.

\vpara{Contributions.}
First, we introduce a state-localized policy-training procedure that combines reward-based filtering with online GRPO restarts from selected partial states, rather than imitating complete accepted outputs. Second, we formulate the downstream verifier as a static-policy value model and use the same signal for search allocation and final selection. Third, we provide controlled evidence beyond the original end-to-end comparisons: full DAPO results, component and curriculum ablations, matched policy--VM pipelines, and approximately latency-matched VM-BoN. Our main empirical evidence remains code reasoning, supplemented by math and science evaluations without target-domain tuning.

\hhide{
To summarize, our contributions are:
\begin{itemize}[leftmargin=*,itemsep=0pt,parsep=0.5em,topsep=0.3em,partopsep=0.3em]
    \item We introduce \model, an LLM RL paradigm that enhances LLM's reasoning ability by \modelTrain and \modelTest.
    The highlight of the method is that it combines the advantages of offline self-training, online learning algorithms, optimized decoding algorithms, and verification methods, achieving balanced efficiency, generalizability, cost, and effectiveness.
    
    \item We show that \model outperforms other reinforcement training baselines (e.g., naive GRPO and ReST-DPO, in Table \ref{tab: policy_training}) and decoding and verification baselines (e.g., PRM-BoN and ORM-MCTS, in Table \ref{tab: reward_model}).
    
    \item Concerning test performance under same training steps, \modelTrain shows higher training efficiency compared to naive GRPO and DAPO (Figure~\ref{fig: main_test} (a)). Given the same budget for decoding verification, \modelTest and the VM achieve better accuracy than Math-Shepherd style PRM or ORM trained on curated public data (Figure~\ref{fig: main_test} (b)).
\end{itemize}
}

\vspace{-0.4cm}
\section{The \model Method}
\label{sec: method}

\begin{figure}[t!]
    \centering
    \includegraphics[width=0.85\linewidth]{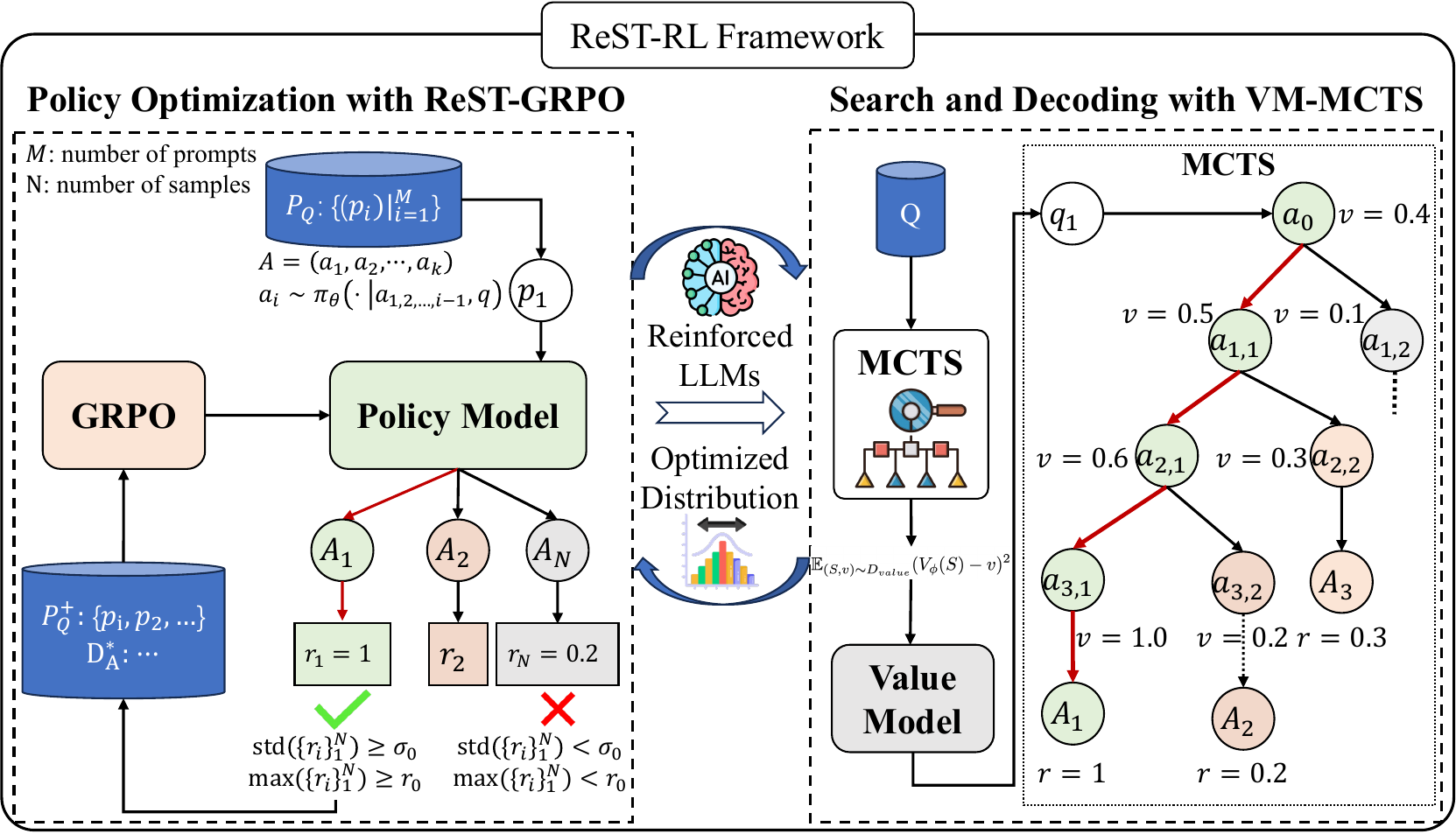}
    \vspace{-0.2cm}
    \caption{
    Framework of \model. \modelTrain reshapes the policy-induced trajectory distribution by filtering and assembling high-value training data. 
    This optimized distribution improves VM learning and enables \modelTest to perform more reliable value-guided search and decoding.
    }
    \label{fig: framework}
\end{figure}

To connect policy optimization and reward learning for LLM reasoning while addressing low-information group rewards and process-level guidance, we develop \model as a unified, state-localized policy--value framework for Reinforced Self-Training.
It is composed of two main components, namely \modelTrain and \modelTest, as shown in Figure \ref{fig: framework}. Importantly, these stages are sequentially connected rather than jointly optimized: \modelTrain first reshapes the trajectory distribution over sparse reward spaces, after which \modelTest estimates values under the resulting static policy. This separation recovers state-level guidance without returning to the full cost and instability of online actor--critic learning. Appendix~\ref{sec: ablation} tests the dependency under matched value learning, and Appendix~\ref{sec: discussion} discusses its practical cost.
\begin{itemize}
[leftmargin=*,itemsep=0pt,parsep=0.5em,topsep=0.3em,partopsep=0.3em]
    \item \textit{\modelTrain} is the policy optimization stage, which integrates ReST-style filtering and partial-state starts into \modelGRPO for sparse-reward reasoning.
    \item \textit{\modelTest} is the reward learning stage, which focuses on specifically training and utilizing a value model (VM) that assists the policy in inference time search and decoding. 
\end{itemize}

\subsection{Task Setup}
\label{sec: task_setup}

For a specific reasoning task, there is generally an instruction or question, denoted as $q$, which describes the background information and problem details in text. There may exist a ground truth answer $g$. For coding tasks, there may also be some test cases $t_{1,2,\dots,m}$. Using an LLM policy known as $\pi_\theta$, we can gradually generate solutions to the problem, following the predicted output token probabilities of the policy. 
This process can be regarded as a Markov Process, where the policy takes an action $a_i$ step by step following $\pi_\theta(a_i|a_{1,2,\dots,i-1},q)$, based on previously generated content $a_1, a_2,\dots,a_{i-1}$ and the instruction $q$. Eventually, when the eos token or stop strings are generated, the process stops, forming a final solution $A=(a_1, a_2,\dots,a_{k})$. 
In our approach, to align with the definitions in common RL, we define an action as a single line of text, which is separated by a line break. In Appendix \ref{sec: discussion}, we further illustrate the reasons for choosing this granularity. We also define an intermediate state as the combination of the instruction and a partial solution that contains lines of text, i.e. $S_i=(q,a_{1,2,\dots,i})$. This means an intermediate state always terminates with a line break. In comparison, an end state stands for the combination of the instruction and a full solution that terminates with eos token, i.e. $S_{end}=S_k=(q,a_{1,2,\dots,k})$.
In consistency with conventional settings for reasoning scenarios, the transition to the next state after an action was taken is considered to be deterministic. 
In terms of reward, we assign rewards only for end states. The reward function for an end state is denoted by $R=R(S_{end})$, which can be either rule-based or model-based. Moreover, the policy value function defined by Eq.~(\ref{eq: value_function}) can be unified with the Q function.
\begin{equation}
    \label{eq: value_function}
    V^\pi(S_i)=\mathbb{E}_\pi[R(S_{end})|S_i],
    Q^\pi(S_i,a_{i+1})=\mathbb{E}_\pi[R(S_{end})|S_{i+1}]=V^\pi(S_{i+1}).
\end{equation}


\subsection{\modelTrain}

Building on ReST \citep{gulcehre2023reinforcedselftrainingrestlanguage} and \modelGRPO \citep{shao2024deepseekmathpushinglimitsmathematical}, Reinforced Self-Training with Group Relative Policy Optimization (\modelTrain) improves \modelGRPO training through reward-based filtering and partial-state starts. Unlike ReST-style supervised training on accepted completions, an accepted prefix is used only as context: its suffix is freshly sampled online and optimized with the GRPO objective. Thus, the stored trajectory determines where online RL may restart but does not prescribe what the policy should generate afterward. The original prompt is also retained whenever it passes filtering, so the assembled set mixes ordinary prompt starts with state-localized starts. This extension reshapes the training distribution and mitigates low-information group rewards \citep{yu2025dapoopensourcellmreinforcement}, while preserving on-policy exploration over newly sampled suffixes. It also increases the policy's coverage of high-reward trajectories used by the subsequent value-learning stage.
\modelTrain is designed to reinforce the policy through multiple self-training iterations, each initialized from the checkpoint produced by the previous round. Every iteration contains three steps: (1) pre-train solution sampling, (2) train-data filtering and assembly, and (3) optimization with the \modelGRPO objective. 
Algorithm~\ref{alg: rest_grpo_algorithm} provides the detailed pseudocode.


\subsubsection{Pre-train Solution Sampling}

Similar to the ReST algorithm that augments the initial training dataset with samples from the policy itself, \modelTrain also utilizes the sample solutions generated from the policy for further train data selection and assembly. Within each iteration, we first collect $N$ solutions for every instruction prompt in the original dataset using the current policy. By configuring the LLM's sampling temperature, we can control the randomness and diversity of these solutions. Then, we utilize a fixed reward function to obtain rewards for all solutions, which are used for subsequent filtering processes.

\subsubsection{Data Filtering by Reward}

An LLM policy's output solutions and their corresponding rewards contain a wealth of information indicating its strength and weakness toward the target domain, which can be leveraged to filter out ineffective training data. For \modelGRPO, policy update relies on group relative advantage. This means effective training signals depend on discrepancies between rewards, offering three key enlightenments: 
\begin{itemize}
[leftmargin=*,itemsep=0pt,parsep=0.5em,topsep=0.3em,partopsep=0.3em]
    \item If a policy's outputs obtain similar rewards on a question, there's nothing much for it to learn from. 
    \item For questions where the current policy faces an enormous action space but only a few output traces lead to a substantial reward, normal sampling from the initial state may be ineffective for training.
    \item For a question that the current policy does not perform well, high-reward solution traces are crucial for training. Since high-reward solutions often share some common patterns, sampling from a partial solution state of a high-reward solution may be helpful for obtaining more high-reward traces, thus beneficial for training.
\end{itemize}

We design our algorithm's filtering and train data assembly process based on these enlightenments. 
Firstly, we use standard deviation to evaluate the variety of rewards. For question prompts that the policy's solutions achieve a reward standard deviation less than a given threshold $\sigma_0$, we filter them out from the train dataset, because they will possibly result in very little improvement of the policy. For other prompts, we add them to the train dataset. 
Secondly, we focus on high-reward solution traces that may be beneficial for training. Specifically, we filter out question prompts that the policy only achieves rewards lower than a certain threshold $r_0$. For other prompts, we extract their corresponding solution trace with the highest reward. Finally, its partial solution states are utilized to assemble new train data, as illustrated in Section \ref{sec: train_data_assembly}. 
Figure~\ref{fig: reward_std} shows that \modelTrain increases reward variance, but this statistic is only a proxy for the informativeness of group-relative updates. We therefore separate it from downstream attribution: Table~\ref{tab: controlled_results}(a) evaluates the policy components on APPS-500 and BCB, while Appendix~\ref{sec: ablation} characterizes the induced trajectory distribution. Table~\ref{tab: controlled_results}(b) then tests whether the first-stage advantage remains after matched value learning and search.

\subsubsection{Train Data Assembly}
\label{sec: train_data_assembly}

Considering that a partial state belonging to a high-reward solution can possibly be a good state for the policy to begin with, we extract a subset of all these partial solution states for new train data assembly. For simplicity, we only select from partial states of the solution trace $A$ with the highest reward for each prompt $p$ (this can be extended to top-k selection). Denoted by $a_{1,2,\dots,j}$, a partial state contains the first $j$ actions (i.e., lines) of $A$ taken by the policy. In the assembly process, we first sample a subset of these partial states $D_A^*$ based on a discrete finite exponential distribution, using a fixed positive exponent factor $\alpha < 1.0$. 
Sampling is performed following the probability mass function defined by Eq.~(\ref{eq: sample_pmf}).
Relative to ordinary \modelGRPO, every selected nonempty prefix moves the starting state along a high-reward trajectory. Within this set of partial states, the distribution below assigns greater mass to shorter, earlier prefixes than uniform sampling, retaining a larger suffix action space for online exploration. We sample $\beta |A|$ partial states and add them to $D_A^*$, where $\beta$ limits the training budget, and append each state to the initial prompt $p$. As $\alpha$ approaches $0$, the mass concentrates on the shortest partial prefix, so the resulting starts approach ordinary GRPO-style starts but are not identical to using only the original prompt.
\begin{equation}
    \label{eq: sample_pmf}
    p(a_{1,2,\dots,j})=\frac{1-\alpha}{1-\alpha^{|A|}}\alpha^{j-1}, j=1,2,\dots,|A|
\end{equation}

\subsubsection{\modelGRPO Training}

After training data for each iteration are obtained, online training is performed. Policy update is based on the objective displayed in Eq.~(\ref{eq: grpo_objective}), which is basically the \modelGRPO objective, but the policy's input prompt can be a combination of question and sampled partial solution from the train set $P_Q^+$.
\begin{equation}
    \label{eq: grpo_objective}
    \begin{aligned}
    \mathcal{J}(\theta)=&\mathbb{E}[p\sim P_Q^+,\{o_i\}_{i=1}^G\sim \pi_{\theta_{old}}(O|p)] \frac{1}{G}\sum_{i=1}^G\frac{1}{|o_i|}\sum_{t=1}^{|o_i|}\{min[\frac{\pi_\theta(o_{i,t}|p,o_{i,<t})}{\pi_{\theta_{old}}(o_{i,t}|p,o_{i,<t})}\hat{A}_{i,t}, \\
    &clip(\frac{\pi_\theta(o_{i,t}|p,o_{i,<t})}{\pi_{\theta_{old}}(o_{i,t}|p,o_{i,<t})},1-\epsilon,1+\epsilon)\hat{A}_{i,t}]-\beta'\mathbb{D}_{KL}[\pi_\theta||\pi_{ref}]\}
    \end{aligned}
\end{equation}

\subsection{\modelTest}

Value Model based Monte-Carlo Tree Search (\modelTest) trains a VM from self-collected MCTS targets under a static policy and uses it for inference-time search and selection. Its target is the expected terminal reward $V^\pi(s)$ of a partial state, not whether the most recent line is independently correct. This distinction is useful in executable-reward settings, where intermediate steps may be difficult to label even though complete solutions can be evaluated automatically. The VM serves two roles: it allocates search toward promising reasoning subspaces and ranks completed candidates. Because $V^\pi$ is policy-dependent, we pair each VM with the policy that generated its training trajectories. The matched comparison in Table~\ref{tab: controlled_results}(b) evaluates this two-stage dependency without introducing policy--VM mismatch. We compare \modelTest with other MCTS-based approaches in Appendix~\ref{sec: discussion}.
\modelTest is inspired by but extends beyond previous process-supervision methods such as Math-Shepherd (M-S) \citep{wang2024mathshepherdverifyreinforcellms}. Rather than serving only as a standalone verifier, the VM is introduced as a value estimator trained on self-collected search data under the static policy, enabling test-time search that is both fine-grained and broadly applicable. Although the value target introduces some extra noise, it leads to better performance when decoding with the assistance of MCTS.

\subsubsection{The Value Target}

In the M-S method, the quality of an intermediate step is defined as its potential to deduce the correct final answer, which is essentially an expectation-based evaluation target. However, this estimation is still treated as a measure of single step correctness. For \modelTest, we instead summarize this to the value target defined in Eq.~(\ref{eq: value_function}), emphasizing the evaluation of the whole partial state including the last action. It naturally reflects the potential for a policy to reach high-reward end states from the current partial state, which can be leveraged to assist policy decoding. In fact, the M-S soft target can be regarded as a special case of value target where the reward function $R$ is defined by Eq.~(\ref{eq: math_shepherd_reward}).
\begin{equation}
\label{eq: math_shepherd_reward}
    R(S_{end})=\mathbb{I}(a_{A}=a^*), \ \ s.t. \ \ S_{end}=(q,A)
\end{equation}

\subsubsection{Value Model Training}

We adopt MCTS when collecting training data for the VM, as it balances exploration of different reasoning paths and exploitation of promising intermediate states, aligning with our design principles. In our algorithm, the root node of the search tree is set to the initial state $S_{init}=S_0=(q,\varnothing)$, while other nodes represent intermediate states or end states $S_i=(q,a_{1,2,\dots,i})$. As illustrated in Section \ref{sec: task_setup}, an action represents a line of text. Through multiple rounds of MC simulation, value targets of partial states and end states are gradually updated and precisely estimated, distinguished from the all-at-once estimation used by M-S. Because these targets are collected under the current static policy, the quality of value learning depends on the policy-induced trajectory distribution, which is precisely why the preceding \modelTrain is crucial. We present the pseudocode for this process in Algorithm \ref{alg: value_train_algorithm}. Once sufficient value data are acquired, a VM $V_\phi$ is trained on these data to predict the value target of different states. The loss function used by \modelTest for training is demonstrated in Eq.~(\ref{eq: value_loss}).
\begin{equation}
\label{eq: value_loss}
    \mathcal{L}_{\phi}=\mathbb{E}_{(S,v)\sim D_{value}}(V_\phi (S)-v)^2
\end{equation} 

\subsubsection{Assisted Decoding}

Previous work such as PPO-MCTS \citep{liu2024dontthrowawayvalue} demonstrates the utility of VMs for token-level MCTS decoding but relies on coupled policy--value training. Here, a separately trained VM predicts expected reward for partial states under a static policy and known rewards for terminal states. As shown in Algorithm~\ref{alg: value_test_algorithm}, the adapted search uses VM estimates for single-step MC rollouts and for final Best-of-N-style selection \citep{lightman2023letsverifystepstep}. Search allocation and final verification therefore use a common state-value scale, rather than combining independently trained guidance and reranking models. Under the assumptions stated in Appendix~\ref{sec: proof}, the value-based rollout remains unbiased and has no greater variance than complete-trace rollout.


\begin{table*}[t!]
  \centering
  \caption{Main results for policy training. LLM policies are evaluated after each training iteration ends. ``Iter.'' refers to sequential training iterations on $Q_\text{train}$ (0th iter. means not trained). The complete DAPO comparison is conducted under the main Qwen3-8B configuration; the remaining model blocks retain the original baseline set.}
  \vspace{-0.2cm}
  \resizebox{0.89\textwidth}{!}{
    \begin{tabular}{cl|cccccc|c}
    \specialrule{.16em}{0pt} {.65ex}
    Model & Training Method & HumanEval & HumanEval+ & MBPP  & MBPP+ & APPS-500 & BCB & Average \\
    \specialrule{.05em}{.4ex}{.65ex}
    \multicolumn{1}{c}{\multirow{10}[10]{*}{Qwen3-8B}} & Base (0th iter.) & 0.829 & 0.780 & 0.717 & 0.622 & 0.118 & 0.418 & 0.503 \\
\cmidrule{2-9}          & \multicolumn{8}{c}{Below are results for sequential training iterations} \\
\cmidrule{2-9}          & ReST-DPO (1st iter.) & 0.854 & 0.799 & 0.730 & 0.627 & 0.152 & 0.434 & 0.523 \\
          & GRPO (1st iter.) & 0.799 & 0.750 & 0.754 & 0.651 & 0.346 & 0.439 & 0.566 \\
          & DAPO (1st iter.) & 0.805 & 0.768 & 0.770 & 0.667 & 0.351 & 0.425 & 0.570 \\
          & \textbf{ReST-GRPO (1st iter.)} & \textbf{0.872} & \textbf{0.817} & \textbf{0.780} & \textbf{0.672} & \textbf{0.377} & \textbf{0.469} & \textbf{0.604} \\
\cmidrule{2-9}          & ReST-DPO (2nd iter.) & 0.841 & \textbf{0.811} & 0.741 & 0.653 & 0.153 & 0.429 & 0.526 \\
          & GRPO (2nd iter.) & 0.829 & 0.787 & 0.757 & 0.667 & 0.403 & 0.436 & 0.590 \\
          & DAPO (2nd iter.) & 0.811 & 0.773 & 0.770 & 0.676 & 0.428 & 0.444 & 0.597 \\
          & \textbf{ReST-GRPO (2nd iter.)} & \textbf{0.860} & 0.805 & \textbf{0.802} & \textbf{0.690} & \textbf{0.565} & \textbf{0.476} & \textbf{0.655} \\
    \specialrule{.05em}{.4ex}{.65ex}
    \multicolumn{1}{c}{\multirow{8}[8]{*}{Qwen2.5-Coder-7B-Instruct}} & Base (0th iter.) & 0.872 & 0.799 & 0.817 & 0.683 & 0.287 & 0.381 & 0.563 \\
\cmidrule{2-9}          & \multicolumn{8}{c}{Below are results for sequential training iterations} \\
\cmidrule{2-9}          & ReST-DPO (1st iter.) & 0.884 & 0.823 & 0.820 & 0.688 & 0.321 & 0.386 & 0.579 \\
          & GRPO (1st iter.) & \textbf{0.896} & \textbf{0.835} & 0.812 & 0.693 & 0.350 & 0.405 & 0.593 \\
          & \textbf{ReST-GRPO (1st iter.)} & 0.872 & 0.817 & 0.823 & 0.701  & 0.401 & 0.441 & 0.612 \\
\cmidrule{2-9}          & ReST-DPO (2nd iter.) & 0.872 & 0.823 & 0.796 & 0.669 & 0.330 & 0.400 & 0.578 \\
          & GRPO (2nd iter.) & 0.872 & 0.799 & 0.815 & 0.685 & 0.356 & 0.403 & 0.586 \\
          & \textbf{ReST-GRPO (2nd iter.)} & 0.890 & \textbf{0.835} & \textbf{0.849} & \textbf{0.712} & \textbf{0.415} & \textbf{0.462} & \textbf{0.630} \\
    \specialrule{.05em}{.4ex}{.65ex}
    \multicolumn{1}{c}{\multirow{8}[8]{*}{DS-Coder-6.7b-Instruct}} & Base (0th iter.) & 0.744 & 0.677 & 0.741 & 0.646 & 0.230  & 0.338 & 0.493 \\
\cmidrule{2-9}          & \multicolumn{8}{c}{Below are results for sequential training iterations} \\
\cmidrule{2-9}          & ReST-DPO (1st iter.) & 0.756 & 0.683 & 0.725 & 0.635 & 0.251 & 0.329 & 0.495 \\
          & GRPO (1st iter.) & 0.726 & 0.646 & 0.735 & 0.632 & 0.262 & 0.342 & 0.493 \\
          & \textbf{ReST-GRPO (1st iter.)} & 0.756 & 0.683 & 0.743 & 0.635 & 0.282 & 0.355 & 0.511 \\
\cmidrule{2-9}          & ReST-DPO (2nd iter.) & 0.756 & 0.689 & 0.720  & 0.632 & 0.242 & 0.329 & 0.492 \\
          & GRPO (2nd iter.) & 0.707 & 0.634 & 0.730  & 0.624 & 0.283 & 0.357 & 0.497 \\
          & \textbf{ReST-GRPO (2nd iter.)} & \textbf{0.793} & \textbf{0.707} & \textbf{0.749} & \textbf{0.646} & \textbf{0.300}   & \textbf{0.368} & \textbf{0.529} \\
    \specialrule{.05em}{.4ex}{.65ex}
    \multicolumn{1}{c}{\multirow{8}[7]{*}{OpenCI-DS-6.7B}} & Base (0th iter.) & 0.756 & 0.707 & 0.722 & 0.630  & 0.204 & 0.331 & 0.486 \\
\cmidrule{2-9}          & \multicolumn{8}{c}{Below are results for sequential training iterations} \\
\cmidrule{2-9}          & ReST-DPO (1st iter.) & 0.756 & 0.677 & 0.714 & 0.622 & 0.212 & 0.351 & 0.487 \\
          & GRPO (1st iter.) & 0.726 & 0.677 & 0.733 & 0.632 & 0.277 & 0.363 & 0.506 \\
          & \textbf{ReST-GRPO (1st iter.)} & 0.726 & 0.677 & 0.725 & 0.635 & 0.281 & 0.374 & 0.509 \\
\cmidrule{2-9}          & ReST-DPO (2nd iter.) & 0.744 & 0.671 & 0.725 & 0.630  & 0.232 & 0.336 & 0.488 \\
          & GRPO (2nd iter.) & 0.750  & 0.689 & \textbf{0.743} & \textbf{0.648} & 0.279 & 0.352 & 0.512 \\
          & \textbf{ReST-GRPO (2nd iter.)} & \textbf{0.774} & \textbf{0.713} & 0.725 & 0.630  & \textbf{0.325} & \textbf{0.377} & \textbf{0.531} \\
          \specialrule{.16em}{.4ex}{0pt}
    \end{tabular}
  }
  \label{tab: policy_training}
\end{table*}

\begin{table*}[t!]
    \centering
    \caption{Average results of \model and different reward verification methods on all benchmarks. 
    All verification is based on $100$ samples. 
    \model is globally optimal when combining the Table \ref{tab: policy_training}.
    }
  \vspace{-0.2cm}
    \resizebox{0.7\textwidth}{!}{
    \begin{tabular}{c|c|c|c|c}
    \specialrule{.16em}{0pt}{.65ex}
    Method & {Qwen3-8B} & {Qwen2.5-Coder-7B-Instruct} & {DS-Coder-6.7b-Instruct} & {OpenCI-DS-6.7B} \\
    \specialrule{.05em}{.4ex}{.65ex}
    Base & 0.503 & 0.563 & 0.493 & 0.486 \\
    \specialrule{.05em}{.4ex}{.65ex}
    ORM & 0.531 & 0.592 & 0.542 & 0.537 \\
    \specialrule{.05em}{.4ex}{.65ex}
    PRM & 0.516 & 0.591 & 0.539 & 0.532 \\
    \specialrule{.05em}{.4ex}{.65ex}
    ORM-MCTS & 0.538 & 0.588 & 0.547 & 0.535 \\
    \specialrule{.05em}{.4ex}{.65ex}
    VM-MCTS & 0.615 & 0.652 & 0.576 & 0.569 \\
    \specialrule{.05em}{.4ex}{.65ex}
    \textbf{ReST-RL} & \textbf{0.689} & \textbf{0.673} & \textbf{0.584} & \textbf{0.583} \\
    \specialrule{.16em}{.4ex}{0pt}
    \end{tabular}
    }
    \label{tab: reward_model}
\end{table*}

\section{Experiments}
\label{sec: experiments}
As an important branch of LLM reasoning tasks, code writing has received much attention in recent years \citep{zhang2023planninglargelanguagemodels,Li_2022,le2022coderlmasteringcodegeneration}. In this research, we also focus on coding problems, considering that they have a more comprehensive test of LLM's reasoning, planning, and invocation abilities. In this section, we present various experiments to validate \model. Since \model jointly involves policy optimization and value-guided decoding, we also include separate evaluation of \modelTrain and \modelTest against their related baselines in order to identify the contribution of each design under controlled settings. These evaluations are used to establish a fair comparison for each stage aside from testing the complete pipeline. As a whole, we further show that the two stages form a coherent dependency chain and that their unity leads to the best end-to-end performance. In addition, to assess practical feasibility from an end-to-end perspective, we provide supplementary analyses of training-time and inference-time cost, latency, and efficiency for \model in Appendix~\ref{sec: discussion}.

\subsection{Experiment Settings}
\label{sec: experiment_settings}


In this section, we present the main experiment settings, including selected LLMs, datasets used for training and base reward function. Additionally, more experimental details for benchmark evaluation are displayed in Appendix \ref{sec: benchmark_details}.

\vpara{Base LLM Policies.}
We primarily test recent code LLMs: Qwen2.5-Coder-7B-Instruct \citep{hui2024qwen2}, CodeQwen1.5-7B-Chat \citep{codeqwen1.5}, Deepseek-Coder-6.7B-Instruct \citep{guo2024deepseekcoderlargelanguagemodel}, and OpenCodeInterpreter-DS-6.7B \citep{zheng2024opencodeinterpreter}. We additionally include Qwen3-8B \citep{qwen3technicalreport}, Llama-3-8B, and Llama-3.1-8B-Instruct \citep{grattafiori2024llama3herdmodels} to test whether the coding improvements extend across both specialized and general base policies.

\vpara{Training Datasets.}
In terms of source datasets for training, we select three well-known open-source coding datasets: BigCodeBench (BCB) \citep{zhuo2025bigcodebenchbenchmarkingcodegeneration}, DS-1000 \citep{lai2022ds1000naturalreliablebenchmark} and APPS \citep{hendrycks2021measuringcodingchallengecompetence}. After filtering out data in the source train splits (or the randomly sampled train split for BCB) without test cases, we obtain a final dataset $Q_\text{train}$ of $6,945$ coding prompts. For experiments mentioned in Section \ref{sec: evaluation_train} and Section \ref{sec: evaluation_decode}, training of policies and VMs with \model and other baselines are all based on this dataset. All APPS-500 results use one immutable random subset, and we will release its exact problem IDs and evaluation script. Appendix~\ref{sec: benchmark_details} specifies the metric and protocol and records that the historical subset-construction seed is unavailable.

\vpara{Base Reward Function.}
According to previous research, the use of model-based learned reward functions in reinforcement learning scenarios may increase the risk of reward hacking \citep{weng2024rewardhack}, causing undesirable effects on our experimental results. Therefore, we primarily adopt a rule-based reward function that uses test cases to calculate reward, as shown in Eq.~(\ref{eq: reward_function}). It evaluates generated codes with the ratio of passed test cases to total test cases, which naturally reflects the functional correctness of these codes. Somewhat differently, two additional rewards have been added to improve stability and effectiveness for \modelTrain and naive \modelGRPO training, which is illustrated in Appendix \ref{sec: training_methods}.
\begin{equation}
    \label{eq: reward_function}
    \resizebox{0.48\textwidth}{!}{$
    \begin{aligned}
    &R_{base}(S_{end})=R_{base}(q,A)=\frac{1}{m}\sum_{i=1}^m \mathbb{I}(\text{eval}(A,t_i)=y_i),
    \end{aligned}
    $}
\end{equation}
$\text{where }t_i\text{ are test cases, } y_i\text{ are desired test outputs.}$

\subsection{Evaluation of LLM Policy Training Methods}
\label{sec: evaluation_train}

To assess \modelTrain, we compare it with \modelDPO (essentially \modelSFT \citep{singh2023beyond} with DPO optimization), naive \modelGRPO, and DAPO \citep{yu2025dapoopensourcellmreinforcement}. These baselines cover offline preference optimization, ordinary online group-relative optimization, and full dynamic sampling. We evaluate this stage separately because its immediate role is to improve the policy and reshape the trajectory distribution before any VM is trained. All policy methods use a learning rate of 1e-7. Appendix~\ref{sec: training_methods} provides their remaining configurations. Table~\ref{tab: controlled_results}(a) reports downstream component controls, while Appendix~\ref{sec: ablation} separates reward-signal diagnostics from final task performance.

\begin{figure*}[t!]
    \centering
    \begin{subfigure}[b]{0.47\textwidth}
        \centering
        \label{fig: training_efficiency_test}
        \includegraphics[height=2.0in,
        valign=t]{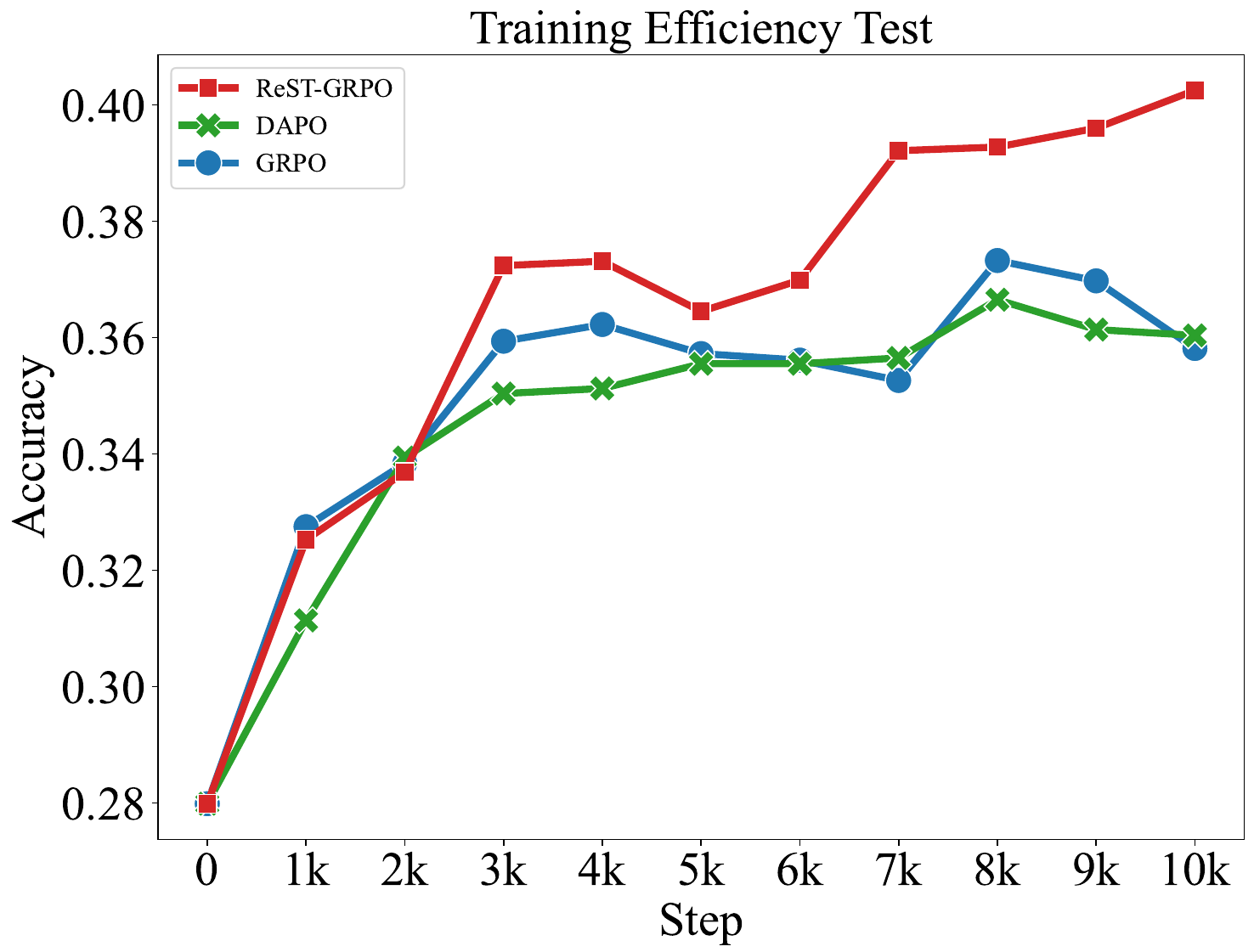}
        \caption{Training efficiency of \modelTrain, DAPO, and \modelGRPO on average benchmark score. 
        Llama-3-8B is trained for $10k$ steps and evaluated every $1k$ steps.}
    \end{subfigure}
    ~
    \begin{subfigure}[b]{0.47\textwidth}
        \centering
        \label{fig: budgeted_verification_test}
        \includegraphics[height=2.0in,
        valign=t]{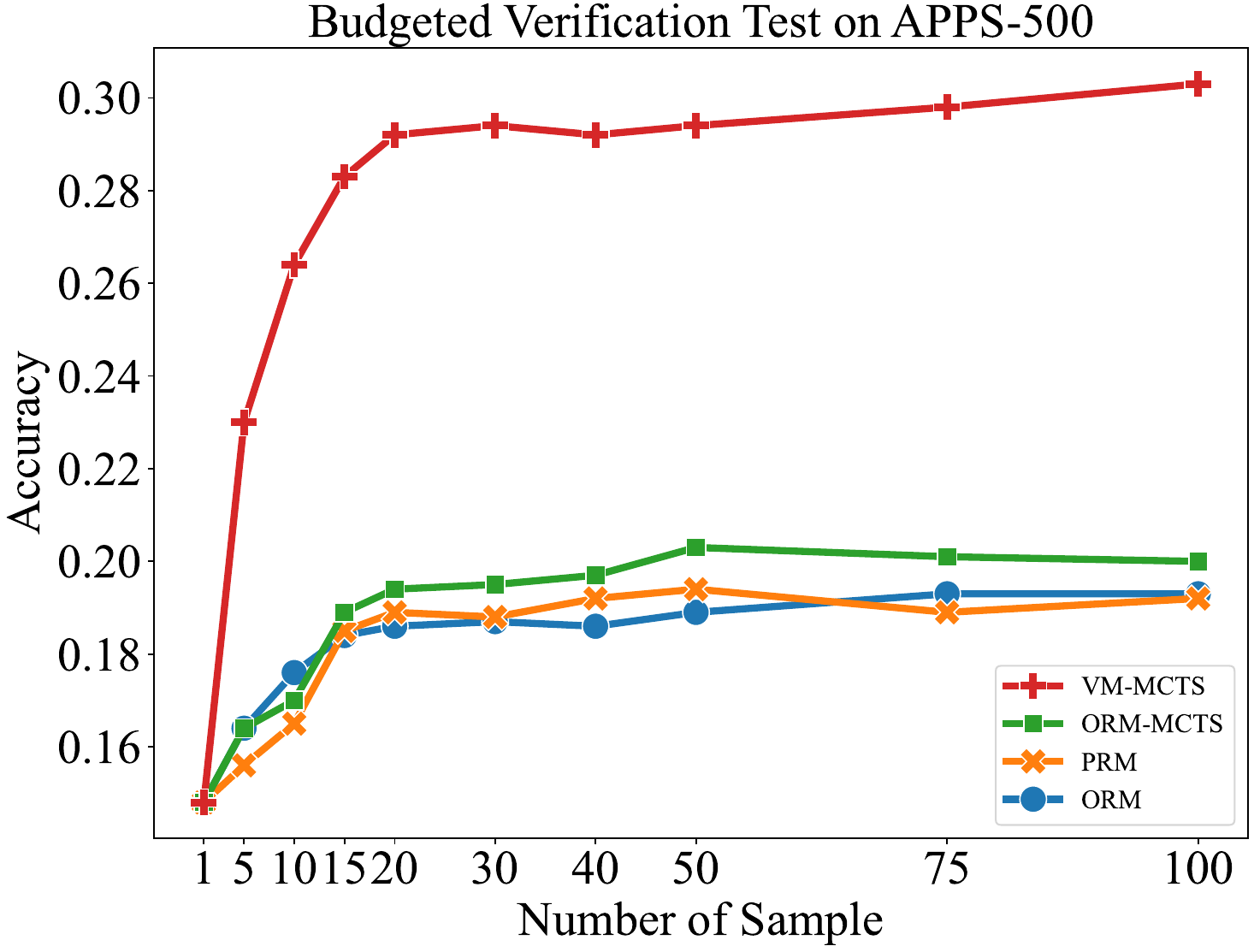}
        \caption{Budgeted verification on APPS-500 with CodeQwen. 
        Sampling temperature is set to $0.7$ for all methods.}
    \end{subfigure}
    \vspace{-0.2cm}
    \caption{Test of training efficiency and budgeted verification.}
    \label{fig: main_test}
\end{figure*}

\vpara{Performance comparison for iterative policy training.}
We perform two sequential training iterations on $Q_\text{train}$. Table~\ref{tab: policy_training} reports four LLM policies, with additional Llama-3.1-8B-Instruct and CodeQwen1.5-7B-Chat results in Appendix~\ref{sec: additional_results}. ReST-GRPO improves the average over ReST-DPO and naive GRPO for every model and iteration. Under the main Qwen3-8B configuration, it also outperforms DAPO on all six benchmarks, with average margins of 3.4 points after the first iteration and 5.8 after the second. After two iterations, ReST-GRPO improves the selected models' average scores by $15.2\%$, $6.7\%$, $3.6\%$, and $4.5\%$, respectively, without degrading any benchmark. Table~\ref{tab: sensitivity_train} shows that the default setting remains competitive under reasonable hyperparameter changes.

\vpara{Assessment of training efficiency for \modelTrain, DAPO and naive \modelGRPO.}
By evaluating policy performance after same number of training steps, we get a glimpse of the difference in training efficiency between \modelTrain, DAPO \citep{yu2025dapoopensourcellmreinforcement} and naive \modelGRPO. 
Figure~\ref{fig: main_test}(a) compares average accuracy for Llama-3-8B over $10k$ steps. We use this general base model because it leaves substantial room for coding improvement and makes differences in optimization efficiency easier to observe than with a heavily code-specialized checkpoint. The three methods are comparable during the first $2k$ steps, after which ReST-GRPO's lead grows. At $10k$ steps its improvement is $12.3\%$, versus $7.8\%$ for naive GRPO and $8.1\%$ for DAPO. Figure~\ref{fig: gpu_hour_test} isolates online training throughput, whereas Table~\ref{tab: e2e_train} includes the one-time offline sampling cost. Under this end-to-end accounting, ReST-GRPO uses 1,752 GPU-hours versus GRPO's 2,080 and reaches a 9\% improvement in 71 versus 207 hours. Thus, the offline stage is amortized by fewer online hours to reach the same performance level.

\vpara{Component and curriculum controls.}
Table~\ref{tab: controlled_results}(a) replaces reward variance as the sole attribution evidence with final downstream results. Filtering-only and partial-only each improve over naive GRPO, but their integration performs best (0.207/0.440 on APPS-500/BCB). E2H-C changes which original prompts are sampled over time and yields 0.189/0.418, whereas ReST-GRPO additionally changes where online RL resumes. Exponential prefix sampling also outperforms uniform sampling in this setting; relative to uniform it favors earlier prefixes and therefore leaves a larger suffix action space. Full settings and end-to-end compute are reported in Appendix~\ref{sec: controlled_ablations}.

\begin{wraptable}{R}{0.49\textwidth}
\vspace{-0.15cm}
\centering
\caption{Controlled Qwen3-8B comparisons. Panel (a) matches optimizer, reward, $G=8$, and 2,000 updates; applicable variants reuse the same offline trajectories. Panel (b) matches each VM to its generating policy. Panel (c) fixes the policy and VM while approximately matching latency.}
\label{tab: controlled_results}
\vspace{0.08cm}
\footnotesize
\setlength{\tabcolsep}{4pt}
\centering
\textbf{(a) Policy components}\par
\vspace{0.08cm}
\begin{tabular}{lcc}
\toprule
Method & APPS-500 & BCB \\
\midrule
Naive GRPO & 0.180 & 0.416 \\
Filtering-only & 0.193 & 0.429 \\
E2H-C & 0.189 & 0.418 \\
Partial-only & 0.190 & 0.425 \\
Uniform-prefix & 0.187 & 0.418 \\
Full ReST-GRPO & \textbf{0.207} & \textbf{0.440} \\
\bottomrule
\end{tabular}

\vspace{0.22cm}
\textbf{(b) Matched policy--VM pipelines}\par
\vspace{0.08cm}
\begin{tabular}{lcc}
\toprule
Pipeline & APPS-500 & BCB \\
\midrule
GRPO (policy only) & 0.403 & 0.436 \\
GRPO + matched VM-MCTS & 0.538 & 0.449 \\
ReST-GRPO (policy only) & 0.565 & 0.476 \\
ReST-GRPO + matched VM-MCTS & \textbf{0.642} & \textbf{0.506} \\
\bottomrule
\end{tabular}

\vspace{0.22cm}
\textbf{(c) Approximately matched latency}\par
\vspace{0.08cm}
\begin{tabular}{lcc}
\toprule
Method & APPS-500 & Time \\
\midrule
VM-BoN ($N=260$) & 0.619 & $\approx$41.5 s$^\dagger$ \\
VM-MCTS ($n=5,T=20$) & \textbf{0.642} & 41.5 s \\
\bottomrule
\end{tabular}
\par\vspace{0.04cm}
{\scriptsize $^\dagger$VM-BoN latency is the calibration target.}
\vspace{-0.2cm}
\end{wraptable}

\subsection{Comparison with Decoding and Verification Methods}
\label{sec: evaluation_decode}

Here we compare \modelTest with representative ORM-, PRM-, and MCTS-based decoding methods. These assessments test whether the learned value signal guides search and verification better once a policy is fixed. We include ORM+BoN (ORM), PRM+BoN (PRM), and \modelORM. Prior work finds MCTS more effective than BFS/DFS tree search in comparable reasoning settings \citep{zhang2024restmctsllmselftrainingprocess,xu2025sramctsselfdrivenreasoningaugmentation}, thus we do not include BFS/DFS based variants as baselines. Appendix~\ref{sec: decoding_methods} provides detailed configurations.
\vpara{Comparison of method accuracy on all benchmarks.}
Table~\ref{tab: reward_model} reports average results with verification limited to 100 candidates. VM-MCTS improves over the strongest listed baseline by $11.2\%$, $8.9\%$, $8.3\%$, and $8.3\%$ for Qwen3-8B, Qwen2.5-Coder, DS-Coder, and OpenCI, respectively. Its lead over \modelORM supports using a state-localized value estimate during decoding rather than only scoring completed outputs. Since all methods use the same candidate budget in this comparison, the result measures sample efficiency. Appendix Table~\ref{tab: sensitivity_test} further shows smooth gains as the branch factor increases.

\vpara{Verification results based on controlled budget.}
Figure~\ref{fig: main_test}(b) compares APPS-500 accuracy across sample budgets with CodeQwen. VM-MCTS remains ahead across the tested range, and Appendix Table~\ref{tab: usage} reports the corresponding token usage. Equal candidate counts, however, do not imply equal latency. We therefore compare the same ReST-GRPO policy and VM at approximately matched mean wall-clock in Table~\ref{tab: controlled_results}(c): VM-MCTS obtains 0.642 versus 0.619 for VM-BoN calibrated at $N=260$ (both about 41.5 s per prompt). Because policy, verifier, and latency are controlled, the remaining difference isolates whether the VM is used during tree allocation or only for final reranking. Appendix Table~\ref{tab: e2e_test} retains the complementary equal-sample latency trade-off.

\begin{table}[H]
  \centering
  \caption{End-to-end training time (hours) assessment for ReST-GRPO and naive GRPO using Llama-3-8B. We conduct training for a total of 10k steps (same number of samples) with each method. Improvement is evaluated based on average score across all benchmarks, and pre-training sampling time for ReST-GRPO is always included.}
  \resizebox{\linewidth}{!}{
    \begin{tabular}{l|cccccc}
    \specialrule{.16em}{0pt}{.65ex}
    Method & Pre-sampling Time & Total Training Time & Total GPU Hours & 6\% Improvement & 9\% Improvement & 12\% Improvement \\
    \specialrule{.05em}{.4ex}{.65ex}
    GRPO & 0 & 260 & 2080 & 52 & 207 & $>$260 \\
    ReST-GRPO & 10 & 219 & 1752 & 51 & 71 & 199 \\
    \specialrule{.16em}{.4ex}{0pt}
    \end{tabular}
  }
  \label{tab: e2e_train}
\end{table}

\subsection{Validation of the Overall Approach}
\label{sec: evaluation_overall}

Based on the final policies trained with \modelTrain, we evaluate the full \model pipeline. Table~\ref{tab: reward_model} reports the best average result for each base model. Appendix Table~\ref{tab: trajectory_distribution} provides mechanism-level evidence: identical MCTS collection yields mean trajectory rewards of 0.37/0.56/0.68 for Base/GRPO/ReST-GRPO. The matched downstream test in Table~\ref{tab: controlled_results}(b) holds VM initialization, architecture, 1.1M targets, collection, optimization, and search fixed. A GRPO-trained VM improves its policy to 0.538/0.449 on APPS-500/BCB, while the matched ReST-GRPO pipeline reaches 0.642/0.506. Thus a learned state-localized VM benefits either policy, and the first-stage policy-induced advantage persists through matched value learning and search.

\begin{table}[H]
  \centering
  \caption{Preliminary cross-domain transfer results with Qwen3-8B. The VM is trained only on code trajectories and receives no target-domain tuning. We also pair the initial VM (0th) with ReST-GRPO (2nd) to assess robustness to policy shift.}
  \resizebox{0.8\linewidth}{!}{
    \begin{tabular}{cl|ccccc}
    \specialrule{.16em}{0pt} {.65ex}
    Model & Method & APPS-500 & BCB & MATH & Omni-MATH & GPQA-Diamond \\
    \specialrule{.05em}{.4ex}{.65ex}
    \multicolumn{1}{c}{\multirow{3}[5]{*}{Qwen3-8B}}
    & Base (0th) & 0.118 & 0.418 & 0.780 & 0.234 & 0.449 \\
    & Base (0th) + VM (0th) & 0.415 & 0.471 & 0.828 & 0.238 & 0.460 \\
    & ReST-GRPO (2nd) + VM (0th) & 0.630 & 0.496 & 0.862 & 0.246 & 0.480 \\
    & ReST-RL (2nd) & \textbf{0.642} & \textbf{0.506} & \textbf{0.872} & \textbf{0.256} & \textbf{0.490} \\
    \specialrule{.16em}{.4ex}{0pt}
    \end{tabular}
  }
  \label{tab: ood}
\end{table}

\vpara{Preliminary cross-domain transfer.}
We investigate whether a code-trained \model policy and VM remain useful on out-of-domain tasks and after policy updates. Note that coding remains the principal empirical domain, and this experiment tests transfer without target-domain tuning rather than claiming a universal verifier.
Table~\ref{tab: ood} reports Qwen3-8B based results on MATH \citep{hendrycks2021measuringmathematicalproblemsolving}, Omni-MATH \citep{gao2024omnimathuniversalolympiadlevel}, and GPQA-Diamond \citep{rein2023gpqagraduatelevelgoogleproofqa}. Without target-domain tuning, adding the code-trained VM to the base policy improves the three scores from 0.780/0.234/0.449 to 0.828/0.238/0.460. The initial VM also remains useful after two policy-training rounds (0.862/0.246/0.480), while updating it to the final shifted policy yields 0.872/0.256/0.490.
These controlled observations provide evidence that the domain-agnostic objective $V^\pi(s)=\mathbb{E}_\pi[R\mid s]$ can transfer across task and policy shifts. The initial VM's retained gains also indicate robustness to moderate policy shift, while the updated matched VM performs best.

\section{Conclusion}
\label{sec: conclusion}

In this paper, we introduce \textbf{\model}, a unified, state-localized policy--value framework for Reinforced Self-Training. \modelTrain combines reward-based filtering with partial-state online RL: selected prefixes determine new starting contexts, while their suffixes are freshly sampled and optimized rather than imitated. \modelTest subsequently learns $V^\pi(s)$ under the resulting static policy and uses this signal for both MCTS allocation and final output selection. The two stages therefore reconnect policy improvement and value-guided reasoning without requiring a jointly trained online actor--critic.

Across multiple coding benchmarks and model families, ReST-GRPO improves over ReST-DPO, naive GRPO, and DAPO, while VM-MCTS improves over representative verification and search baselines. Controlled component, matched policy--VM, and approximately latency-matched comparisons further support the distinct contribution of state-localized training and value-guided allocation. End-to-end accounting shows that the offline sampling cost can be amortized by more efficient online training. Finally, math and science results without target-domain tuning provide transfer evidence beyond our primary coding domain.

\section*{Acknowledgements}
This work was supported by Fundamental and Interdisciplinary Disciplines Breakthrough Plan of the Ministry of Education of China (No. JYB2025XDXM101), NSFC for Distinguished Young Scholar (62425601), New Cornerstone Science Foundation through the XPLORER PRIZE and CIPS-SMP-Zhipu Large Model Fund (No. CIPS-SMP20250319).

\bibliography{NIPS2026-CameraReady/nips2026}
\bibliographystyle{NIPS2026-CameraReady/nips2026}

\clearpage
\appendix

\section{Appendix}

\subsection{Ablation and Sensitivity Analysis}
\label{sec: ablation}

We analyze \model from both training and inference perspectives. The results address two questions: whether performance is robust to reasonable hyperparameter changes, and whether the policy-induced trajectory shift remains consequential after matched value learning. Downstream component controls are separated from the reward-variance proxy to avoid treating an optimization statistic as final task performance.

\vpara{\modelTrain: reshaping distribution for an appropriate search policy.}
Table~\ref{tab: ablation} is a reward-standard-deviation proxy analysis: filtering increases mean SD from 0.104 to 0.148, while high-reward trajectory selection and partial-state sampling further increase it to 0.168. Final downstream attribution is reported separately in Table~\ref{tab: component_ablation}. Table~\ref{tab: sensitivity_train} varies $\sigma_0$, $\alpha$, and $r_0$ and shows a relatively narrow performance range, with the default configuration strongest overall. These results indicate robustness to reasonable perturbations without implying that reward variance alone establishes task-level gains.

\vpara{\modelTest: unlock the potential of search under controlled budgets.}
Table~\ref{tab: sensitivity_test} studies the branch factor $n$ in \modelTest. APPS-500 accuracy increases from the policy-only score of 0.565 to 0.652 as the search becomes more expressive. Gains are already substantial at the main setting $n=5$, indicating a smooth compute--performance trade-off rather than dependence on one branch factor.

\vpara{Policy-induced trajectories and matched value learning.}
Table~\ref{tab: trajectory_distribution} uses the same MCTS collection procedure as VM training. Relative to Base and GRPO, ReST-GRPO increases mean trajectory reward from 0.37/0.56 to 0.68 and median reward from 0.01/0.60 to 0.78. This establishes a distributional shift, while Table~\ref{tab: controlled_results}(b) tests its downstream consequence. With VM initialization, architecture, 1.1M targets, collection parameters, optimization steps, seed, and VM-MCTS fixed, the matched GRPO and ReST-GRPO pipelines obtain 0.538/0.449 and 0.642/0.506 on APPS-500/BCB, respectively. Because a VM estimates $V^\pi$, each VM is evaluated with the policy that generated its trajectories; a mismatched VM would mix data quality with policy--VM distribution shift.

\vpara{Practical suggestions for hyperparameter selection.}
The ablations also suggest practical defaults. With rewards in $[0,1]$, $\sigma_0$ can start in a small range such as $(0,0.1)$ to avoid discarding informative prompts. In $p(j)\propto\alpha^{j-1}$, lower $\alpha$ concentrates mass on shorter prefixes, while $\alpha$ near one approaches a uniform distribution over prefix lengths; $\alpha=0.95$ provides a mild early-prefix bias in our setting. For $r_0$, stricter thresholds favor more reliable high-reward anchors. For \modelTest, $n$ and $T$ should be selected according to the available latency budget.

\vpara{Overall implication.}
Together, the sensitivity, trajectory, and matched-pipeline results show that the method is stable under reasonable hyperparameter changes and that its two stages remain complementary under controlled value training and search.

\begin{table}[H]
  \centering
  \caption{Trajectory reward statistics for Qwen3-8B and its trained variants under the same MCTS collection procedure used for VM training. ReST-GRPO shifts the sampled distribution toward higher-reward regions; the downstream consequence is evaluated directly in Table~\ref{tab: controlled_results}(b).}
  \resizebox{0.45\linewidth}{!}{
    \begin{tabular}{c|ccc}
    \specialrule{.16em}{0pt} {.65ex}
    Method & Base & GRPO & ReST-GRPO \\
    \specialrule{.05em}{.4ex}{.65ex}
    Mean Reward & 0.37 & 0.56 & 0.68 \\
    Median Reward & 0.01 & 0.60 & 0.78 \\
    \specialrule{.16em}{.4ex}{0pt}
    \end{tabular}
  }
  \label{tab: trajectory_distribution}
\end{table}

\begin{table}[H]
    \centering
    \caption{Reward-standard-deviation proxy analysis for train-data sampling variants. This table characterizes the training signal; downstream component performance is reported separately in Table~\ref{tab: component_ablation}.}
    \resizebox{0.7\textwidth}{!}{
    \begin{tabular}{c|c|c|c|c|c}
    \specialrule{.16em}{0pt}{.65ex}
    Sampling Method & {$\sigma_0=0,\beta=0$} & {$\alpha \rightarrow 0(0.01)$} & {$r_0=0.1$} & {$r_0=0.5$} & {$r_0=0.9$} \\
    \specialrule{.05em}{.4ex}{.65ex}
    Mean SD & 0.104 & 0.148 & 0.157 & 0.163 & \textbf{0.168} \\
    \specialrule{.05em}{.4ex}{.65ex}
    Median SD & 0.000 & 0.120 & 0.111 & 0.124 & \textbf{0.130} \\
    \specialrule{.16em}{.4ex}{0pt}
    \end{tabular}
    }
    \label{tab: ablation}
\end{table}

\begin{table}[t!]
    \centering
    \caption{Average token usage per prompt for CodeQwen. We compare ORM + Best-of-$N$ with our \modelTest method, showing the number of decoded tokens under various sampling sizes.}
    \resizebox{0.65\textwidth}{!}{
        \begin{tabular}{lllllll}
        \specialrule{.16em}{0pt}{.65ex}
        Model                     & Method    & $N$=5  & $N$=10 & $N$=20 & $N$=50  & $N$=100 \\
        \specialrule{.05em}{.4ex}{.65ex}
        \multirow{2}{*}{CodeQwen} & Best-of-$N$   & 1290 & 2566 & 5137 & 12800 & 25564 \\
        \cmidrule{2-7}
                                  & \textbf{VM-MCTS} & 1287 & 2544 & 4962 & 12055 & 22427 \\
        \specialrule{.16em}{.4ex}{0pt}
        \end{tabular}
    }
    \label{tab: usage}
\end{table}

\begin{table}[t!]
  \centering
  \caption{Sensitivity analysis for the main hyperparameters of ReST-GRPO. Qwen3-8B is trained with each setting for 1k steps, and the influence of each hyperparameter is assessed individually.}
  \resizebox{0.48\linewidth}{!}{
    \begin{tabular}{c|ccc}
    \specialrule{.16em}{0pt}{.65ex}
    Hyperparameter & Value 1 & Value 2 & Value 3 \\
    \specialrule{.05em}{.4ex}{.65ex}
    $\sigma_0$ & 0 & 0.05 & 0.2 \\
    Acc (APPS) & 0.139 & 0.140 & 0.133 \\
    $\alpha$ & 0.01 & 0.5 & 0.95 \\
    Acc (APPS) & 0.131 & 0.137 & 0.140 \\
    $r_0$ & 0.1 & 0.5 & 0.9 \\
    Acc (APPS) & 0.134 & 0.138 & 0.140 \\
    \specialrule{.16em}{.4ex}{0pt}
    \end{tabular}
  }
  \label{tab: sensitivity_train}
\end{table}

\begin{table}[t!]
  \centering
  \caption{Sensitivity to the VM-MCTS branch factor on APPS-500. The policy is Qwen3-8B after two ReST-GRPO iterations; evaluation uses $T=20$ and otherwise follows the main decoding configuration.}
  \resizebox{0.65\linewidth}{!}{
    \begin{tabular}{c|ccccc}
    \specialrule{.16em}{0pt}{.65ex}
    $n$ (branch factor) & None (Policy only) & 2 & 3 & 5 & 8 \\
    \specialrule{.05em}{.4ex}{.65ex}
    APPS-500 ($T=20$) & 0.565 & 0.613 & 0.625 & 0.642 & 0.652 \\
    \specialrule{.16em}{.4ex}{0pt}
    \end{tabular}
  }
  \label{tab: sensitivity_test}
\end{table}

\subsection{Controlled Component and Compute Ablations}
\label{sec: controlled_ablations}

Table~\ref{tab: component_ablation} reports final downstream performance from the same Qwen3-8B checkpoint. Filtering-only and Partial-only improve APPS-500 over naive GRPO by 1.3 and 1.0 points, while Full ReST-GRPO improves by 2.7 points and also performs best on BCB. Uniform-prefix changes only the prefix-position distribution and falls below the exponential rule. The latter favors earlier prefixes relative to uniform sampling, leaving more of the suffix available for online exploration. Relative to ordinary GRPO, both rules still introduce nonempty partial-state starts.

E2H-C follows the cosine easy-to-hard schedule of \citet{parashar2025curriculum}. It forms four quantile bins using empirical pass rate from the same stored trajectories and gradually shifts sampling from easier to harder original prompts. It never uses partial-state starts. Its 0.189/0.418 APPS-500/BCB result shows that prompt-level scheduling alone does not match state-localized training in this controlled setting.

\begin{table}[t!]
\centering
\caption{Downstream component ablations from the same Qwen3-8B checkpoint. Each variant uses the same reward, optimizer, rollout group $G=8$, and 2,000 updates; variants requiring offline statistics reuse the same $N=30$ trajectories. GPU-hours in this table cover online training only; end-to-end accounting is in Table~\ref{tab: training_compute_qwen}.}
\resizebox{\linewidth}{!}{
\begin{tabular}{lccccl}
\toprule
Method & Prompt selection & Partial starts & Prefix rule & APPS-500 / BCB & Online GPU-h \\
\midrule
Naive GRPO & None & No & -- & 0.180 / 0.416 & 410 \\
Filtering-only & Static filter & No & -- & 0.193 / 0.429 & 434 \\
E2H-C & Dynamic E2H & No & -- & 0.189 / 0.418 & 406 \\
Partial-only & None & Yes & Exponential & 0.190 / 0.425 & 370 \\
Uniform-prefix & Static filter & Yes & Uniform & 0.187 / 0.418 & 294 \\
Full ReST-GRPO & Static filter & Yes & Exponential & \textbf{0.207 / 0.440} & 309 \\
\bottomrule
\end{tabular}}
\label{tab: component_ablation}
\end{table}

Table~\ref{tab: training_compute_qwen} distinguishes online training cost from the shared offline trajectory cost. The 309 GPU-hours reported for Full ReST-GRPO's controlled run are online GPU-hours, including 96 offline GPU-hours gives 405 total, compared with 410 for naive GRPO. We likewise separate offline and online tokens and reward calls rather than presenting the online portion as end-to-end cost.

\begin{table*}[t!]
\centering
\caption{End-to-end compute accounting for the 2,000-update Qwen3-8B controlled runs. Offline trajectories are generated once and reused by all applicable variants; totals nevertheless include their attributable offline cost. Reward calls count solution-level executions.}
\resizebox{0.92\textwidth}{!}{
\begin{tabular}{l|rrr|rrr|rrr|r}
\toprule
& \multicolumn{3}{c|}{GPU-hours} & \multicolumn{3}{c|}{Sampled tokens (M)} & \multicolumn{3}{c|}{Reward calls} & \\
Method & Offline & Online & Total & Offline & Online & Total & Offline & Online & Total & Updates \\
\midrule
Naive GRPO & 0 & 410 & 410 & 0 & 22.4 & 22.4 & 0 & 32k & 32k & 2,000 \\
Filtering-only & 96 & 434 & 530 & 56.3 & 22.3 & 78.6 & 208k & 32k & 240k & 2,000 \\
E2H-C & 96 & 406 & 502 & 56.3 & 22.4 & 78.7 & 208k & 32k & 240k & 2,000 \\
Partial-only & 96 & 370 & 466 & 56.3 & 29.2 & 85.5 & 208k & 32k & 240k & 2,000 \\
Uniform-prefix & 96 & 294 & 390 & 56.3 & 23.8 & 80.1 & 208k & 32k & 240k & 2,000 \\
Full ReST-GRPO & 96 & 309 & \textbf{405} & 56.3 & 24.1 & 80.4 & 208k & 32k & 240k & 2,000 \\
\bottomrule
\end{tabular}}
\label{tab: training_compute_qwen}
\end{table*}

\subsection{Benchmark Evaluation Details}
\label{sec: benchmark_details}

We evaluate on nine benchmarks: HumanEval \citep{chen2021evaluatinglargelanguagemodels}, HumanEval+ \citep{liu2023codegeneratedchatgptreally}, MBPP \citep{austin2021programsynthesislargelanguage}, MBPP+ \citep{liu2023codegeneratedchatgptreally}, BigCodeBench (BCB) \citep{zhuo2025bigcodebenchbenchmarkingcodegeneration}, APPS \citep{hendrycks2021measuringcodingchallengecompetence}, MATH \citep{hendrycks2021measuringmathematicalproblemsolving}, Omni-MATH \citep{gao2024omnimathuniversalolympiadlevel}, and GPQA-Diamond \citep{rein2023gpqagraduatelevelgoogleproofqa}. We report pass@1 except on APPS-500, where we follow APPS and average the fraction of passed test cases. All evaluations are zero-shot.

\vpara{APPS-500 reproducibility.}
All experiments use the same immutable 500-problem subset stored as \texttt{data/APPS/test\_500.jsonl} and no response experiment resamples it. We will release this file together with problem IDs/indices, evaluation scripts and commands, prompt templates, metric definitions, temperature, and maximum generation length. The historical random seed used to construct the subset was not logged, so we do not claim that the sampling process can be reconstructed from a seed. Releasing the exact subset instead fully determines the evaluated examples.

\subsection{Main Algorithms}
\label{sec: main_algorithms}
We present the self-training \modelTrain algorithm in Algorithm~\ref{alg: rest_grpo_algorithm}, the automated training data collection process of VM in Algorithm~\ref{alg: value_train_algorithm} and the assisted decoding algorithm of \modelTest in Algorithm~\ref{alg: value_test_algorithm}.

\begin{algorithm}[ht!]
\small
\caption{\modelTrain algorithm for policy training.}
\renewcommand{\algorithmicrequire}{\textbf{Input:}}
\renewcommand{\algorithmicensure}{\textbf{Output:}}
\label{alg: rest_grpo_algorithm}
\begin{algorithmic}[1]
\REQUIRE {base LLM policy $\pi_\theta$, reward function (model) $R$, original question dataset $Q$, original prompt dataset $P_Q$, number of solution samples $N$, standard deviation threshold $\sigma_0$, reward threshold $r_0$, ratio of data to sample $\beta$, sampling exponent factor $\alpha$, number of iterations $T$.}
\FOR {$i=1$ to $T$}
    \STATE $P_Q^+ \leftarrow \varnothing$ 
    \pycomment{initialize train set}
    \FOR {prompt $p$ in $P_Q$}
        \STATE Sample $N$ solutions $\{A_i\}_{i=1}^N \sim \pi_\theta(\cdotp|p)$ 
        \pycomment{generate solutions samples}
        \STATE Get reward $\{r_i=R(p,A_i)\}_{i=1}^N$ 
        \pycomment{obtain rewards for samples}
        \STATE $\sigma \leftarrow std(\{r_i\}_{i=1}^N)$ 
        \pycomment{calculate standard deviation for reward}
        \IF {$\sigma \geq \sigma_0$}
            \STATE $P_Q^+ \leftarrow p$
            \IF {$max_i(r_i) \geq r_0$}
                \STATE $A \leftarrow argmax_{A_i}(r_i)$ 
                \pycomment{get the solution with highest reward}
                \STATE $D_A \leftarrow \{a_{1,2,\dots,j}\}_{j=1}^{|A|}$ 
                \pycomment{extract partial solutions}
                \STATE $D_A^* \leftarrow do\_sample(D_A,\beta,\alpha)$ 
                \pycomment{sample partial solutions for train data assembly}
                \FOR {partial solution $p^*$ in $D_A^*$}
                    \STATE $P_Q^+ \leftarrow p + p^*$
                \ENDFOR
            \ENDIF
        \ENDIF
    \ENDFOR
    \STATE $\pi_\theta \leftarrow GRPO(P_Q^+,\pi_\theta,R)$ 
    \pycomment{train the policy with the GRPO objective}
\ENDFOR
\ENSURE $\pi_\theta$
\end{algorithmic}
\end{algorithm}

\begin{algorithm}[t!]
\caption{The value model's train data collection process.}
\renewcommand{\algorithmicrequire}{\textbf{Input:}}
\renewcommand{\algorithmicensure}{\textbf{Output:}}
\label{alg: value_train_algorithm}
\begin{algorithmic}[1]
\REQUIRE {
LLM policy $\pi_\theta$, 
reward function (model) $R$, 
question (instruction) dataset $Q$, 
prompt dataset $P_Q$,
max search iterations $T$,
number of samples for each expansion $n$, 
exploration constant $c$,
small constant that avoids zero-division $\epsilon$,
initial value $v_0$,
}
\STATE $D_{value}\leftarrow \varnothing$ 
\pycomment{initialize value train set}
\FOR {prompt $p$ in $P_Q$}
    \STATE Initialize the search tree $T_p$
    \FOR {$t=1,2,\dots,T$}
        \STATE $S\leftarrow S_{init}$
        \STATE \textit{\textbf{---------------Node Selection---------------}}
        \WHILE {$S$ is expanded}
            \STATE $S\leftarrow argmax_{S^{'}\in \text{children}(S)}(v_{S^{'}}+c\sqrt{\frac{\ln{N_S}+1}{N_{S^{'}}+\epsilon}})$  
            \pycomment{select child node based on UCT}
        \ENDWHILE
        \STATE $v,m\leftarrow 0$ 
        \pycomment{for value calculation in  backpropagation process}
        \IF {$S=(p,a_{1,2,\dots,j})$ is not an end state}
            \STATE \textit{\textbf{---------------Node Expansion---------------}}
            \STATE Sample $n$ traces $\{o_i=(a_{j+1,j+2,\dots}^{(i)})\}_{i=1}^n \sim \pi_\theta(O|S)$ starting from $S$
            \STATE Build child nodes by taking actions $\{a_{j+1}^{(i)}\}_{i=1}^n$ 
            \pycomment{use the first action of sample traces to expand}
            \FOR {node $S^{'}$ in children$(S)$}
                \STATE $v_{S^{'}}\leftarrow v_0$
                \STATE $N_{S^{'}},U_{S^{'}}\leftarrow 0$
            \ENDFOR
            \STATE \textit{\textbf{---------------MC Rollout---------------}}
            \STATE Get rewards with reward function $\{r_i=R(p,a_{1,2,\dots,j},o_i)\}_{i=1}^n$
            \STATE $D_{value}\leftarrow \{((p,a_{1,2,\dots,j},o_i),r_i)\}_{i=1}^n$ 
            \pycomment{add value targets for end states to train set}
            \STATE $v,m\leftarrow sum(\{r_i\}_{i=1}^n),n$
        \ELSE
            \STATE $r\leftarrow R(S)$ 
            \pycomment{get the reward for current end state}
            \STATE $v,m\leftarrow r,1$
        \ENDIF
        \STATE \textit{\textbf{---------------Value Backpropagation---------------}}
        \FOR {node $S^{'}$ from $S$ to root}
            \STATE $N_{S^{'}}\leftarrow N_{S^{'}}+m$ 
            \pycomment{update visit count}
            \STATE $U_{S^{'}}\leftarrow U_{S^{'}}+v$ 
            \pycomment{update sum of reward}
            \STATE $v_{S^{'}}\leftarrow \frac{U_{S^{'}}}{N_{S^{'}}}$ 
            \pycomment{update value estimation}
        \ENDFOR
    \ENDFOR
    \FOR {node $S$ in $T_p$}
        \IF {$S$ is expanded}
            \STATE $D_{value}\leftarrow (S,v_S)$ 
            \pycomment{add value targets for partial states to train set}
        \ENDIF
    \ENDFOR
\ENDFOR
\STATE \textbf{Return} $D_{value}$
\ENSURE $D_{value}$
\end{algorithmic}
\end{algorithm}

\begin{algorithm}[ht!]
\caption{The assisted decoding algorithm of \modelTest.}
\renewcommand{\algorithmicrequire}{\textbf{Input:}}
\renewcommand{\algorithmicensure}{\textbf{Output:}}
\label{alg: value_test_algorithm}
\begin{algorithmic}[1]
\REQUIRE {
LLM policy $\pi_\theta$, 
Value model $V_\phi$,
original question (instruction) $q$, 
question prompt $p_q$,
max search iterations $T$,
number of samples for each expansion $n$, 
exploration constant $c$,
small constant that avoids zero-division $\epsilon$,
}
\STATE Initialize the search tree $T_{p_q}$
\STATE $D_{sol}\leftarrow \varnothing$ 
\pycomment{initialize solution set}
\FOR {$t=1,2,\dots,T$}
    \STATE $S\leftarrow S_{init}$
    \STATE \textit{\textbf{---------------Node Selection---------------}}
    \WHILE {$S$ is expanded}
        \STATE $S\leftarrow argmax_{S^{'}\in \text{children}(S)}(v_{S^{'}}+c\sqrt{\frac{\ln{N_S}+1}{N_{S^{'}}+\epsilon}})$  
        \pycomment{select child node based on UCT}
    \ENDWHILE
    \STATE $v,m\leftarrow 0$ 
    \pycomment{for value calculation in  backpropagation process}
    \IF {$S=(p,a_{1,2,\dots,j})$ is not an end state}
        \STATE \textit{\textbf{---------------Node Expansion---------------}}
        \STATE Sample $n$ traces $\{o_i=(a_{j+1,j+2,\dots}^{(i)})\}_{i=1}^n \sim \pi_\theta(O|S)$ starting from $S$
        \STATE $D_{sol}\leftarrow \{(a_{1,2,\dots,j},o_i)\}_{i=1}^n$
        \pycomment{add sample solutions to solution set}
        \STATE Build child nodes by taking actions $\{a_{j+1}^{(i)}\}_{i=1}^n$ 
        \pycomment{use the first action of sample traces to expand}
        \STATE \textit{\textbf{---------------Value Based MC Rollout---------------}}
        \FOR {node $S^{'}$ in children$(S)$}
            \STATE $v_{S^{'}}\leftarrow V_\phi(S^{'})$ 
            \pycomment{get value estimation}
            \STATE Set $N_{S^{'}}$ to the number of times $S^{'}$ was visited during expansion phase
            \STATE $U_{S^{'}}\leftarrow N_{S^{'}}v_{S^{'}}$
            \STATE $v\leftarrow v+U_{S^{'}}$
        \ENDFOR
        \STATE $m\leftarrow n$
    \ELSE
        \STATE $v,m\leftarrow v_S,1$
    \ENDIF
    \STATE \textit{\textbf{---------------Value Backpropagation---------------}}
    \FOR {node $S^{'}$ from $S$ to root}
        \STATE $N_{S^{'}}\leftarrow N_{S^{'}}+m$ 
        \pycomment{update visit count}
        \STATE $U_{S^{'}}\leftarrow U_{S^{'}}+v$ 
        \pycomment{update sum of reward}
        \STATE $v_{S^{'}}\leftarrow \frac{U_{S^{'}}}{N_{S^{'}}}$ 
        \pycomment{update value estimation}
    \ENDFOR 
\ENDFOR
\STATE Get value (reward) estimation of every solution $A_i\in D_{sol}$: $\{r_i=V_\phi(p_q,A_i)\}_{i=1}^{|D_{sol}|}$ 
\STATE $A^*=argmax_{A_i}(r_i)$ 
\pycomment{Best-of-N}
\STATE \textbf{Return} $A^*$
\ENSURE $A^*$
\end{algorithmic}
\end{algorithm}

\subsection{Details for Compared Policy Training Methods}
\label{sec: training_methods}
\begin{itemize}
[leftmargin=*,itemsep=0pt,parsep=0.5em,topsep=0.3em,partopsep=0.3em]
    \item \textbf{\modelTrain:} In addition to test-case pass rate, the training reward checks required output substrings and penalizes redundant characters, as shown in Eq.~(\ref{eq: reward_function_grpo}). We set $\omega_1=1e-3$, $\omega_2=1e-6$, $N=30$, $\sigma_0=0.05$, $r_0=0.9$, $\beta=0.5$, $\alpha=0.95$, and rollout group $G=8$. The original prompt is retained when it passes filtering; sampled nonempty prefixes are added as further online-GRPO starts.
    \item \textbf{\modelGRPO:} To validate the effectiveness of the proposed partial sampling methodology on \modelGRPO, we involve naive \modelGRPO as a baseline. For this baseline, the same reward function $R_\text{GRPO}$ is adopted for training. The number of generations per group is also set to $8$. In contrast, different from \modelTrain, questions in $Q_\text{train}$ are directly used for training. This means that each question in $Q_\text{train}$ will be used once within a single training iteration.
    \item \textbf{DAPO:} We use the same Qwen3-8B initialization and training data as the main policy comparison and implement DAPO dynamic sampling, which resamples and removes groups with zero reward variance. Following the DAPO configuration, we use $\epsilon_\text{high}=0.28$, the decoupled GRPO loss, and mask truncated completions. Other shared optimization and rollout settings follow the main comparison.
    \item \textbf{E2H-C:} This controlled curriculum baseline follows the cosine schedule of E2H Reasoner \citep{parashar2025curriculum}. Empirical pass rate from the shared $N=30$ trajectories defines four quantile difficulty bins. Sampling gradually shifts from easy to hard original prompts over 2,000 updates; no partial state is used, while reward, optimizer, and $G=8$ match the component runs.
    \item \textbf{\modelDPO:} We compare another ReST baseline \modelDPO, which is similar to the \modelSFT \citep{singh2023beyond} method, except that the policy update algorithm is replaced from SFT to DPO \citep{rafailov2024direct}. For \modelDPO, we also perform solution sampling with an initial LLM policy on $Q_\text{train}$ within each self-train iteration, with the number of samples per question set to $30$. Subsequently, the base reward function $R_\text{base}$ is deployed to obtain rewards for generated solutions, which are then used for the construction of preference pairs. Finally, we update the policy with the DPO objective, based on generated preference data.
\end{itemize}
\begin{equation}
    \label{eq: reward_function_grpo}
    \begin{aligned}
    &R_\text{GRPO}(q,A)=R_\text{base}(q,A)+\omega_1*\mathbb{I}(s\subseteq A)-\omega_2*n\_redundant\_char(A), \\
    &\text{where }s\text{ is an essential substring, and }\omega_1,\omega_2\text{ are weight parameters.}
    \end{aligned}
\end{equation}

\subsection{Details for Compared Decoding and Verification Methods}
\label{sec: decoding_methods}
We provide detailed descriptions and configurations of compared decoding and verification methods, i.e., \modelTest, ORM, PRM, and \modelORM.
\begin{itemize}
[leftmargin=*,itemsep=0pt,parsep=0.5em,topsep=0.3em,partopsep=0.3em]
    \item \textbf{\modelTest:} For each policy, we train a corresponding VM with Algorithm~\ref{alg: value_train_algorithm} on $Q_\text{train}$. Value-data collection uses $T=30$, $n=5$, $c=0.4$, $\epsilon=0.1$, and $v_0=0$; the VM is initialized by adding a classifier head to the policy network. Test-time decoding uses $c=0.1$. Main ReST-RL results and the matched controls use $T=20,n=5$; other budget experiments vary $T$ as stated.
    \item \textbf{ORM:} We use the Skywork-Reward-Llama-3.1-8B-v0.2 ORM, which is trained on high-quality preference pairs sourced from publicly available data \citep{liu2024skywork}. The output with the highest reward is returned.
    \item \textbf{PRM:} A Qwen2.5 PRM trained with an improved M-S method \citep{zhang2025lessonsdevelopingprocessreward} is adopted to provide action-level reward. We use the minimum action-level reward for a single output as the final verification score. The output with the highest score is returned.
    \item \textbf{\modelORM:} To assess the influence of MCTS on decoding accuracy, we include this baseline that also uses the Skywork ORM. We deploy a MCTS decoding algorithm similar to Algorithm \ref{alg: value_test_algorithm}, except that value estimation and final verification is based on the ORM.
\end{itemize}

\subsection{Additional Experimental Results}
\label{sec: additional_results}

\vpara{Additional results for policy training.} 
Different from training results in Table~\ref{tab: policy_training}, we provide policy training results when using greedy decoding in test in Table~\ref{tab: policy_training_greedy}. In addition, we also provide the training results of Llama-3.1-8B-Instruct and CodeQwen1.5-7B-Chat in Table~\ref{tab: policy_training_other}. To ensure that the improvements achieved are beyond noise, we conduct four independent replicate self-train experiments under identical conditions mentioned in Section \ref{sec: evaluation_train}. In Table \ref{tab: policy_training_sd}, we report self-train results with std-dev for Qwen3-8B on every benchmark. It can be observed that \modelTrain maintains clear margins ($>5\%$ Avg., clearly not within random noise) over all other baselines and across iterations, indicating its significant improvement.

\begin{table}[t!]
  \centering
  \caption{Policy training results when using greedy decoding in test.}
  \resizebox{\textwidth}{!}{
    \begin{tabular}{cl|cccccc|c}
    \specialrule{.16em}{0pt} {.65ex}
    Model & Training Method & HumanEval & HumanEval+ & MBPP  & MBPP+ & APPS-500 & BCB & Avg. \\
    \specialrule{.05em}{.4ex}{.65ex}
    \multicolumn{1}{c}{\multirow{8}[8]{*}{Qwen3-8B}} & Base (0th iter.) & 0.848 & 0.787 & 0.717 & 0.619 & 0.116 & 0.430 & 0.508 \\
\cmidrule{2-9}          & \multicolumn{8}{c}{Below are results for sequential training iterations} \\
\cmidrule{2-9}          & ReST-DPO (1st iter.) & 0.817 & 0.762 & 0.730 & 0.638 & 0.140 & 0.407 & 0.505 \\
          & GRPO (1st iter.) & 0.829 & 0.793 & 0.743 & 0.640 & 0.347 & 0.445 & 0.574 \\
          & \textbf{ReST-GRPO (1st iter.)} & 0.841 & 0.799 & 0.775 & 0.664 & 0.405 & 0.468 & 0.603 \\
\cmidrule{2-9}          & ReST-DPO (2nd iter.) & 0.823 & 0.774 & 0.741 & 0.656 & 0.153 & 0.416 & 0.517 \\
          & GRPO (2nd iter.) & 0.829 & 0.787 & 0.757 & 0.656 & 0.412 & 0.448 & 0.594 \\
          & \textbf{ReST-GRPO (2nd iter.)} & 0.878 & 0.829 & 0.794 & 0.698 & 0.552 & 0.482 & \textbf{0.658} \\
    \specialrule{.05em}{.4ex}{.65ex}
    \multicolumn{1}{c}{\multirow{8}[8]{*}{Qwen2.5-Coder-7B-Instruct}} & Base (0th iter.) & 0.909 & 0.841 & 0.828 & 0.714 & 0.302 & 0.419 & 0.592 \\
\cmidrule{2-9}          & \multicolumn{8}{c}{Below are results for sequential training iterations} \\
\cmidrule{2-9}          & ReST-DPO (1st iter.) & 0.909 & 0.841 & 0.831 & 0.714 & 0.340 & 0.416 & 0.601 \\
          & GRPO (1st iter.) & 0.896 & 0.835 & 0.833 & 0.714 & 0.384 & 0.425 & 0.612 \\
          & \textbf{ReST-GRPO (1st iter.)} & 0.896 & 0.848 & 0.860 & 0.730 & 0.436 & 0.468 & 0.643 \\
\cmidrule{2-9}          & ReST-DPO (2nd iter.) & 0.909 & 0.848 & 0.825 & 0.714 & 0.356 & 0.422 & 0.607 \\
          & GRPO (2nd iter.) & 0.890 & 0.829 & 0.844 & 0.728 & 0.392 & 0.428 & 0.616 \\
          & \textbf{ReST-GRPO (2nd iter.)} & 0.909 & 0.860 & 0.854 & 0.725 & 0.416 & 0.492 & \textbf{0.646} \\
    \specialrule{.05em}{.4ex}{.65ex}
    \multicolumn{1}{c}{\multirow{8}[8]{*}{DS-Coder-6.7b-Instruct}} & Base (0th iter.) & 0.768 & 0.695 & 0.751 & 0.659 & 0.242 & 0.346 & 0.506 \\
\cmidrule{2-9}          & \multicolumn{8}{c}{Below are results for sequential training iterations} \\
\cmidrule{2-9}          & ReST-DPO (1st iter.) & 0.780  & 0.707 & 0.746 & 0.661 & 0.245 & 0.348 & 0.510 \\
          & GRPO (1st iter.) & 0.780  & 0.713 & 0.765 & 0.677 & 0.277 & 0.350  & 0.524 \\
          & \textbf{ReST-GRPO (1st iter.)} & 0.787 & 0.707 & 0.767 & 0.685 & 0.287 & 0.366 & 0.532 \\
\cmidrule{2-9}          & ReST-DPO (2nd iter.) & 0.780  & 0.707 & 0.746 & 0.659 & 0.254 & 0.350  & 0.513 \\
          & GRPO (2nd iter.) & 0.762 & 0.683 & 0.759 & 0.675 & 0.286 & 0.356 & 0.520 \\
          & \textbf{ReST-GRPO (2nd iter.)} & 0.780  & 0.701 & 0.767 & 0.672 & 0.323 & 0.390  & \textbf{0.543} \\
    \specialrule{.05em}{.4ex}{.65ex}
    \multicolumn{1}{c}{\multirow{8}[7]{*}{OpenCI-DS-6.7B}} & Base (0th iter.) & 0.787 & 0.713 & 0.738 & 0.648 & 0.210  & 0.365 & 0.505 \\
\cmidrule{2-9}          & \multicolumn{8}{c}{Below are results for sequential training iterations} \\
\cmidrule{2-9}          & ReST-DPO (1st iter.) & 0.774 & 0.713 & 0.733 & 0.643 & 0.207 & 0.373 & 0.503 \\
          & GRPO (1st iter.) & 0.780  & 0.713 & 0.735 & 0.656 & 0.312 & 0.365 & 0.530 \\
          & \textbf{ReST-GRPO (1st iter.)} & 0.780  & 0.713 & 0.738 & 0.656 & 0.324 & 0.373 & 0.535 \\
\cmidrule{2-9}          & ReST-DPO (2nd iter.) & 0.768 & 0.701 & 0.738 & 0.646 & 0.218 & 0.360  & 0.501 \\
          & GRPO (2nd iter.) & 0.780  & 0.720  & 0.728 & 0.643 & 0.308 & 0.361 & 0.526 \\
          & \textbf{ReST-GRPO (2nd iter.)} & 0.774 & 0.713 & 0.749 & 0.664 & 0.324 & 0.387 & \textbf{0.540} \\
          \specialrule{.16em}{.4ex}{0pt}
    \end{tabular}
  }
  \label{tab: policy_training_greedy}
\end{table}

\begin{table}[t!]
  \centering
  \caption{Policy training results for Llama-3.1-8B-Instruct and CodeQwen1.5-7B-Chat.}
  \resizebox{\textwidth}{!}{
    \begin{tabular}{cl|cccccc|c}
    \specialrule{.16em}{0pt} {.65ex}
    Model & Training Method & HumanEval & HumanEval+ & MBPP  & MBPP+ & APPS-500 & BCB & Avg. \\
    \specialrule{.05em}{.4ex}{.65ex}
    \multicolumn{1}{c}{\multirow{16}[7]{*}{Llama-3.1-8B-Instruct}} & Base & 0.671 & 0.591 & 0.693 & 0.569 & 0.175 & 0.301 & 0.435 \\
          & Base (greedy) & 0.689 & 0.610 & 0.664 & 0.545 & 0.213 & 0.325 & 0.448 \\
\cmidrule{2-9}          & \multicolumn{8}{c}{Below are results for the first training iteration} \\
\cmidrule{2-9}          & ReST-DPO & 0.628 & 0.579 & 0.669 & 0.569 & 0.209 & 0.313 & 0.436 \\
          & GRPO & 0.652 & 0.573 & 0.675 & 0.566 & 0.271 & 0.324 & 0.457 \\
          & \textbf{ReST-GRPO} & 0.677 & 0.610  & 0.735 & 0.616 & 0.248 & 0.383 & \textbf{0.488} \\
\cmidrule{2-9}          & ReST-DPO (greedy) & 0.677 & 0.616 & 0.701 & 0.574 & 0.259 & 0.321 & 0.466 \\
          & GRPO (greedy) & 0.683 & 0.634 & 0.706 & 0.579 & 0.329 & 0.329 & 0.490 \\
          & \textbf{ReST-GRPO (greedy)} & 0.707 & 0.646 & 0.725 & 0.608 & 0.312 & 0.407 & \textbf{0.516} \\
\cmidrule{2-9}          & \multicolumn{8}{c}{Below are results for the second training iteration} \\
\cmidrule{2-9}          & ReST-DPO & 0.671 & 0.598 & 0.672 & 0.556 & 0.249 & 0.299 & 0.449 \\
          & GRPO & 0.671 & 0.598 & 0.701 & 0.585 & 0.284 & 0.341 & 0.476 \\
          & \textbf{ReST-GRPO} & 0.665 & 0.598 & 0.693 & 0.587 & 0.235 & 0.418 & \textbf{0.481} \\
\cmidrule{2-9}          & ReST-DPO (greedy) & 0.677 & 0.616 & 0.698 & 0.574 & 0.272 & 0.325 & 0.470 \\
          & GRPO (greedy) & 0.701 & 0.634 & 0.696 & 0.563 & 0.307 & 0.343 & 0.487 \\
          & \textbf{ReST-GRPO (greedy)} & 0.720 & 0.659 & 0.741 & 0.624 & 0.342 & 0.425 & \textbf{0.535} \\

    \specialrule{.05em}{.4ex}{.65ex}
    \multicolumn{1}{c}{\multirow{16}[7]{*}{CodeQwen1.5-7B-Chat}} & Base & 0.805 & 0.732 & 0.810  & 0.677 & 0.148 & 0.348 & 0.502 \\
          & Base (greedy) & 0.854 & 0.787 & 0.831 & 0.706 & 0.163 & 0.374 & 0.532 \\
\cmidrule{2-9}          & \multicolumn{8}{c}{Below are results for the first training iteration} \\
\cmidrule{2-9}          & ReST-DPO & 0.823 & 0.774 & 0.759 & 0.659 & 0.167 & 0.353 & 0.507 \\
          & GRPO & 0.829 & 0.750  & 0.788 & 0.653 & 0.244 & 0.365 & 0.530 \\
          & \textbf{ReST-GRPO} & 0.854 & 0.793 & 0.804 & 0.690 & 0.270 & 0.389 & \textbf{0.557} \\
\cmidrule{2-9}          & ReST-DPO (greedy) & 0.854 & 0.787 & 0.817 & 0.696 & 0.185 & 0.373 & 0.534 \\
          & GRPO (greedy) & 0.866 & 0.799 & 0.817 & 0.698 & 0.272 & 0.401 & 0.566 \\
          & \textbf{ReST-GRPO (greedy)} & 0.890  & 0.835 & 0.831 & 0.714 & 0.294 & 0.447 & \textbf{0.594} \\
\cmidrule{2-9}          & \multicolumn{8}{c}{Below are results for the second training iteration} \\
\cmidrule{2-9}          & ReST-DPO & 0.841 & 0.780  & 0.799 & 0.672 & 0.176 & 0.339 & 0.515 \\
          & GRPO & 0.878 & 0.823 & 0.815 & 0.672 & 0.255 & 0.375 & 0.556 \\
          & \textbf{ReST-GRPO} & 0.878 & 0.816 & 0.828 & 0.706 & 0.278 & 0.425 & \textbf{0.579} \\
\cmidrule{2-9}          & ReST-DPO (greedy) & 0.860  & 0.787 & 0.836 & 0.709 & 0.211 & 0.362 & 0.542 \\
          & GRPO (greedy) & 0.866 & 0.799 & 0.828 & 0.701 & 0.296 & 0.401 & 0.574 \\
          & \textbf{ReST-GRPO (greedy)} & 0.902 & 0.835 & 0.836 & 0.725 & 0.308 & 0.479 & \textbf{0.609} \\
    \specialrule{.16em}{.4ex}{0pt}
    \end{tabular}
  }
  \label{tab: policy_training_other}
\end{table}

\begin{table}[t!]
  \centering
  \caption{Reinforce self training results for Qwen3-8B with standard deviation.}
  \resizebox{\textwidth}{!}{
    \begin{tabular}{cl|cccccc|c}
    \specialrule{.16em}{0pt} {.65ex}
    Model & Training Method & HumanEval & HumanEval+ & MBPP  & MBPP+ & APPS-500 & BCB & Avg. \\
    \specialrule{.05em}{.4ex}{.65ex}
    \multicolumn{1}{c}{\multirow{8}[8]{*}{Qwen3-8B}} & Base (0th iter.) & 0.829±0.018 & 0.780±0.019 & 0.717±0.003 & 0.622±0.002 & 0.118±0.003 & 0.418±0.007 & 0.503±0.008 \\
\cmidrule{2-9}          & \multicolumn{8}{c}{Below are results for sequential training iterations} \\
\cmidrule{2-9}          & ReST-DPO (1st iter.) & 0.854±0.013 & 0.799±0.006 & 0.730±0.005 & 0.627±0.011 & 0.152±0.015 & 0.434±0.010 & 0.523±0.011 \\
          & GRPO (1st iter.) & 0.799±0.018 & 0.750±0.030 & 0.754±0.016 & 0.651±0.013 & 0.346±0.006 & 0.439±0.004 & 0.566±0.012 \\
          & \textbf{ReST-GRPO (1st iter.)} & 0.872±0.012 & 0.817±0.006 & 0.780±0.005 & 0.672±0.003 & 0.377±0.008 & 0.469±0.013 & 0.604±0.009 \\
\cmidrule{2-9}          & ReST-DPO (2nd iter.) & 0.841±0.006 & 0.811±0.012 & 0.741±0.003 & 0.653±0.007 & 0.153±0.011 & 0.429±0.007 & 0.526±0.008 \\
          & GRPO (2nd iter.) & 0.829±0.006 & 0.787±0.006 & 0.757±0.006 & 0.667±0.014 & 0.403±0.007 & 0.436±0.004 & 0.590±0.007 \\
          & \textbf{ReST-GRPO (2nd iter.)} & 0.860±0.012 & 0.805±0.012 & 0.802±0.006 & 0.690±0.010 & 0.565±0.022 & 0.476±0.007 & \textbf{0.655}±0.012 \\
          \specialrule{.16em}{.4ex}{0pt}
    \end{tabular}
  }
  \label{tab: policy_training_sd}
\end{table}


\begin{figure}[t!]
    \centering
    \includegraphics[height=2.0in]{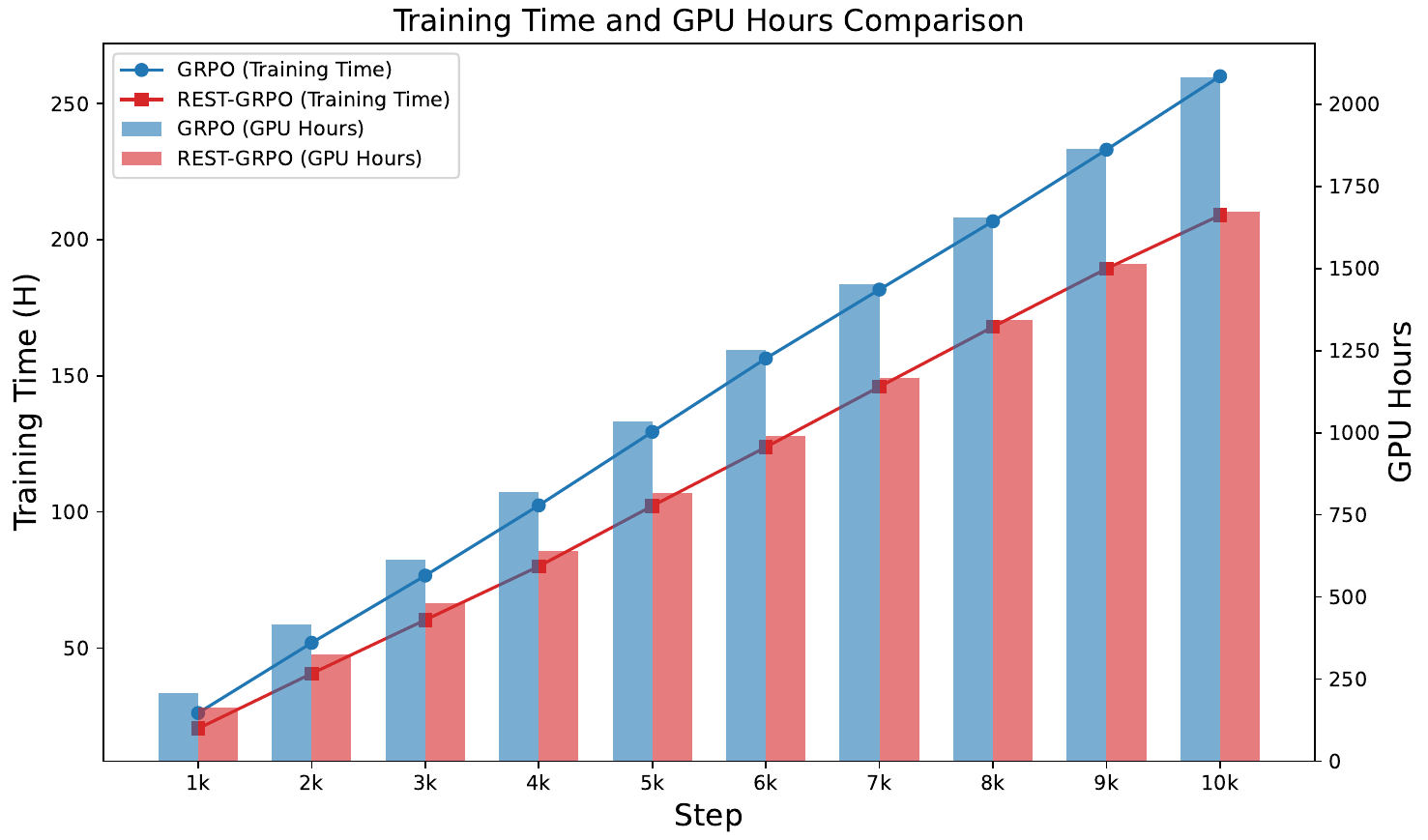}
    \caption{Training time and corresponding GPU-hour accounting for Llama-3-8B, comparing \modelTrain with naive GRPO. The training time is recorded every $1000$ steps. Note that the time for pre-train sampling of \modelTrain is not included. See Table~\ref{tab: e2e_train} for a full end-to-end training time comparison.}
    \label{fig: gpu_hour_test}
\end{figure}

\begin{figure}[t!]
    \centering
    \begin{subfigure}[b]{0.485\textwidth}
        \centering
        \label{fig: CodeQwen}
        \includegraphics[height=1.9in]{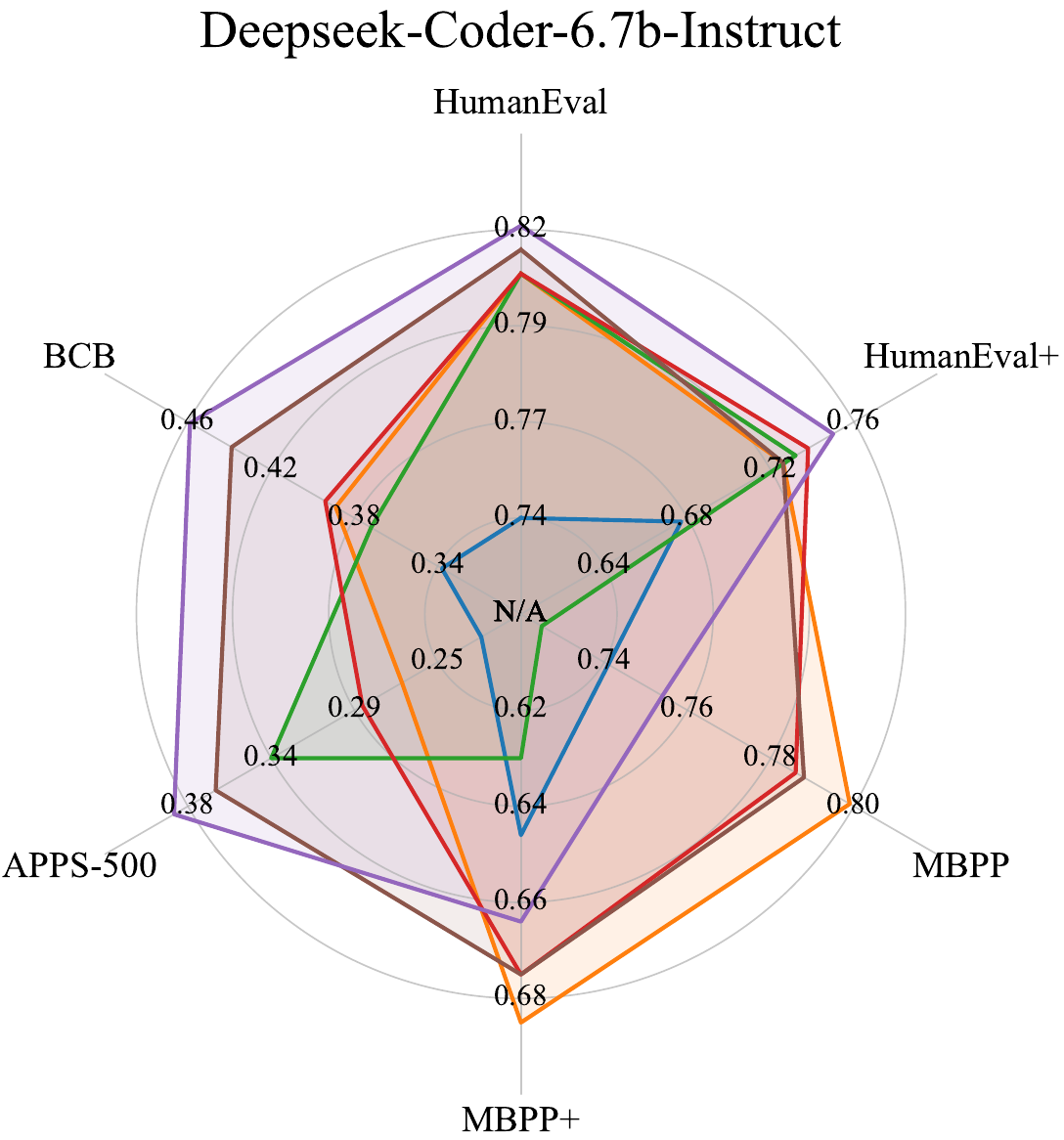}
    \end{subfigure}
    ~
    \begin{subfigure}[b]{0.485\textwidth}
        \centering
        \label{fig: Qwen3}
        \includegraphics[height=1.9in]{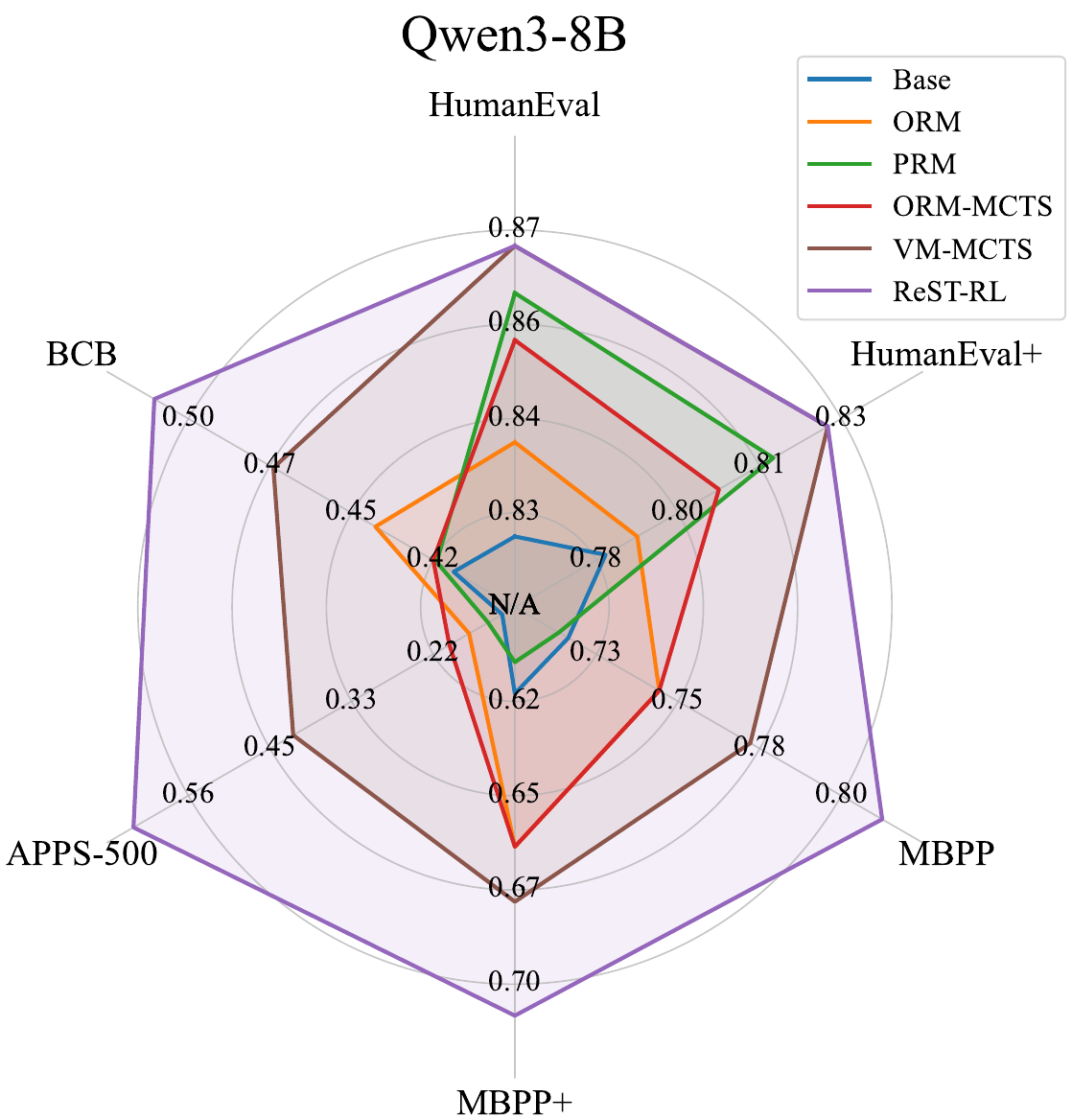}
    \end{subfigure}
    \vskip\baselineskip
    \begin{subfigure}[b]{0.485\textwidth}
        \centering
        \label{fig: OCI-DS}
        \includegraphics[height=1.9in]{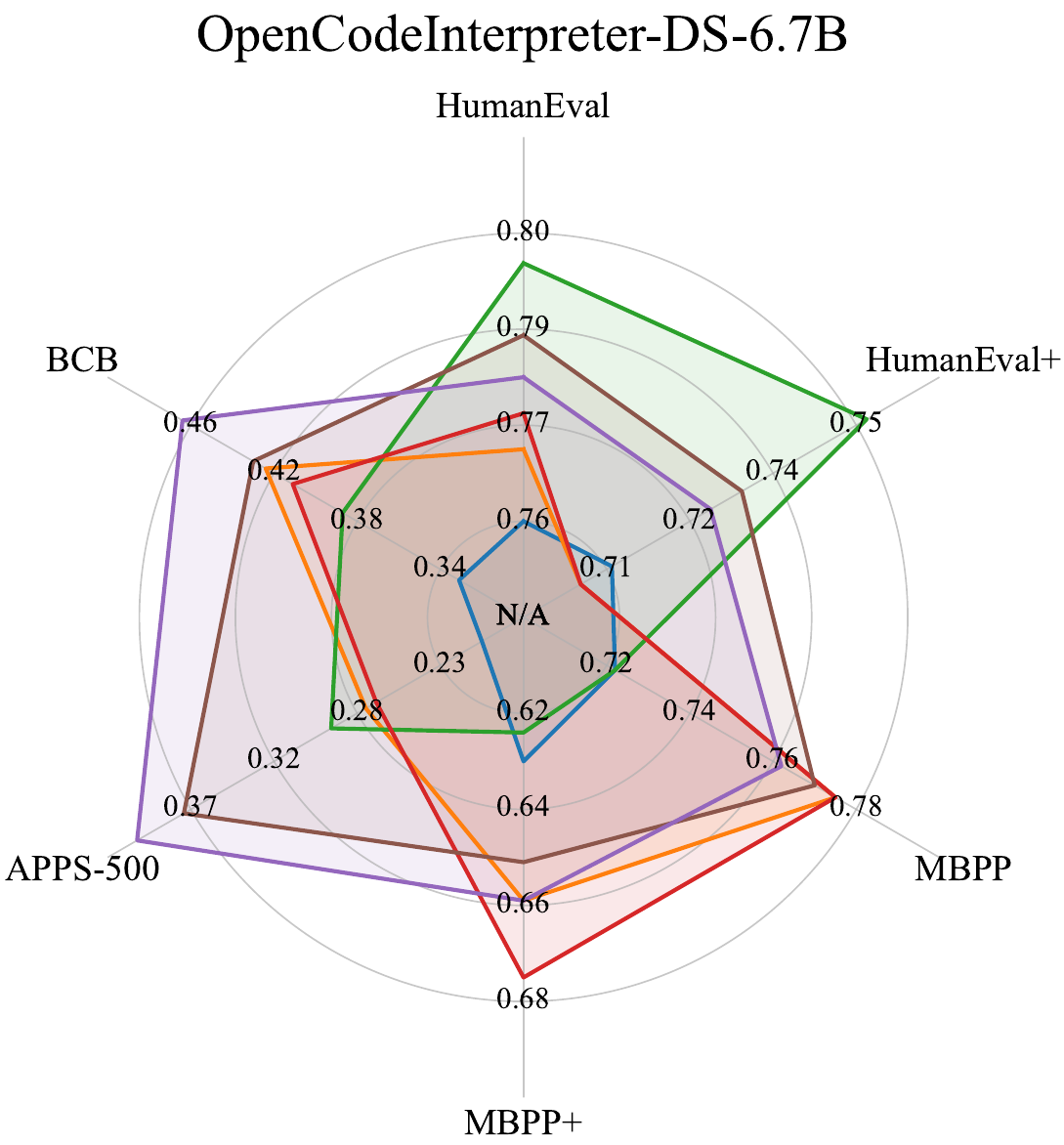}
    \end{subfigure}
    \hfill
    \begin{subfigure}[b]{0.485\textwidth}
        \centering
        \label{fig: Qwen2.5-Code}
        \includegraphics[height=1.9in]{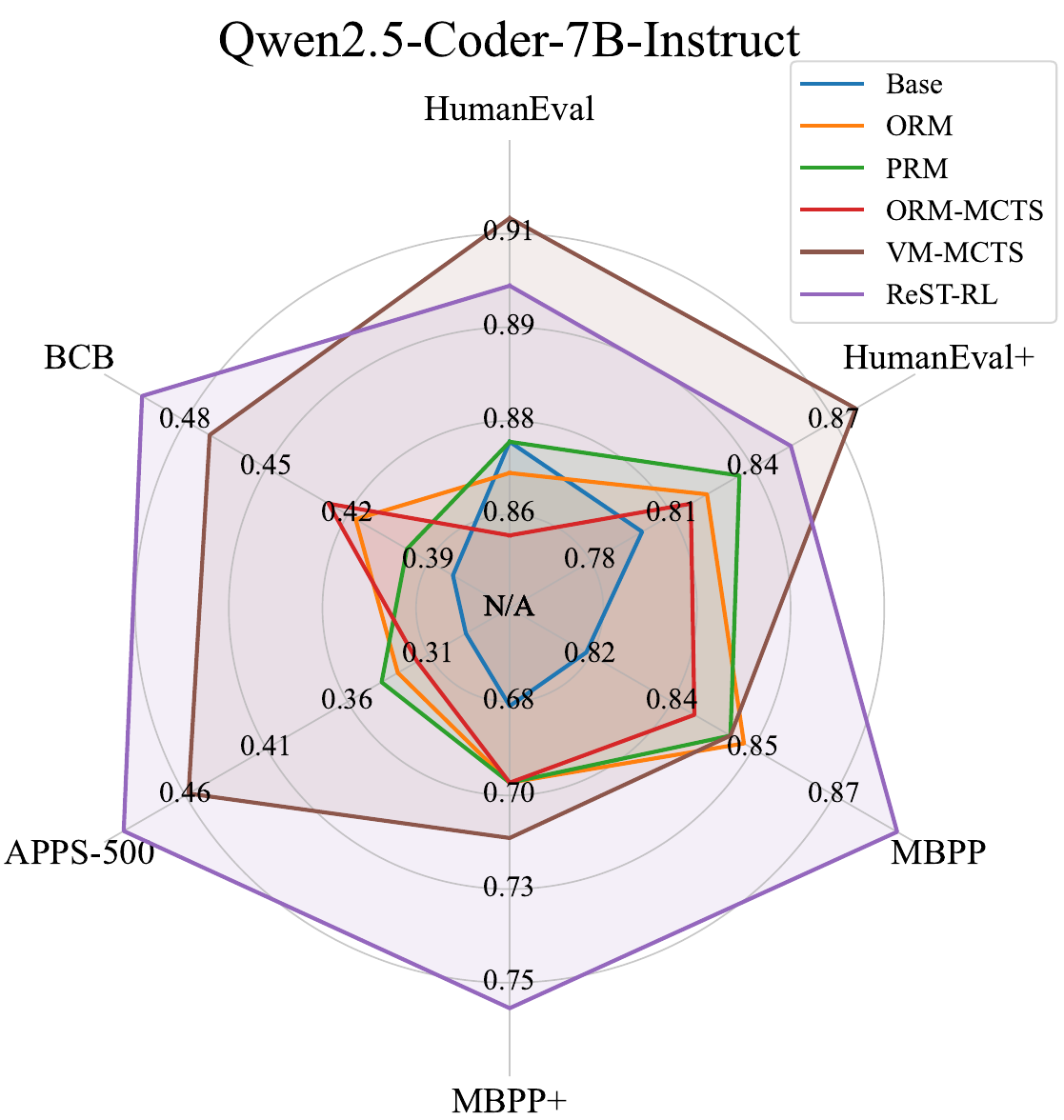}
    \end{subfigure}
    \vskip\baselineskip
    \begin{subfigure}[b]{0.485\textwidth}
        \centering
        \label{fig: Llama-3.1}
        \includegraphics[height=1.9in]{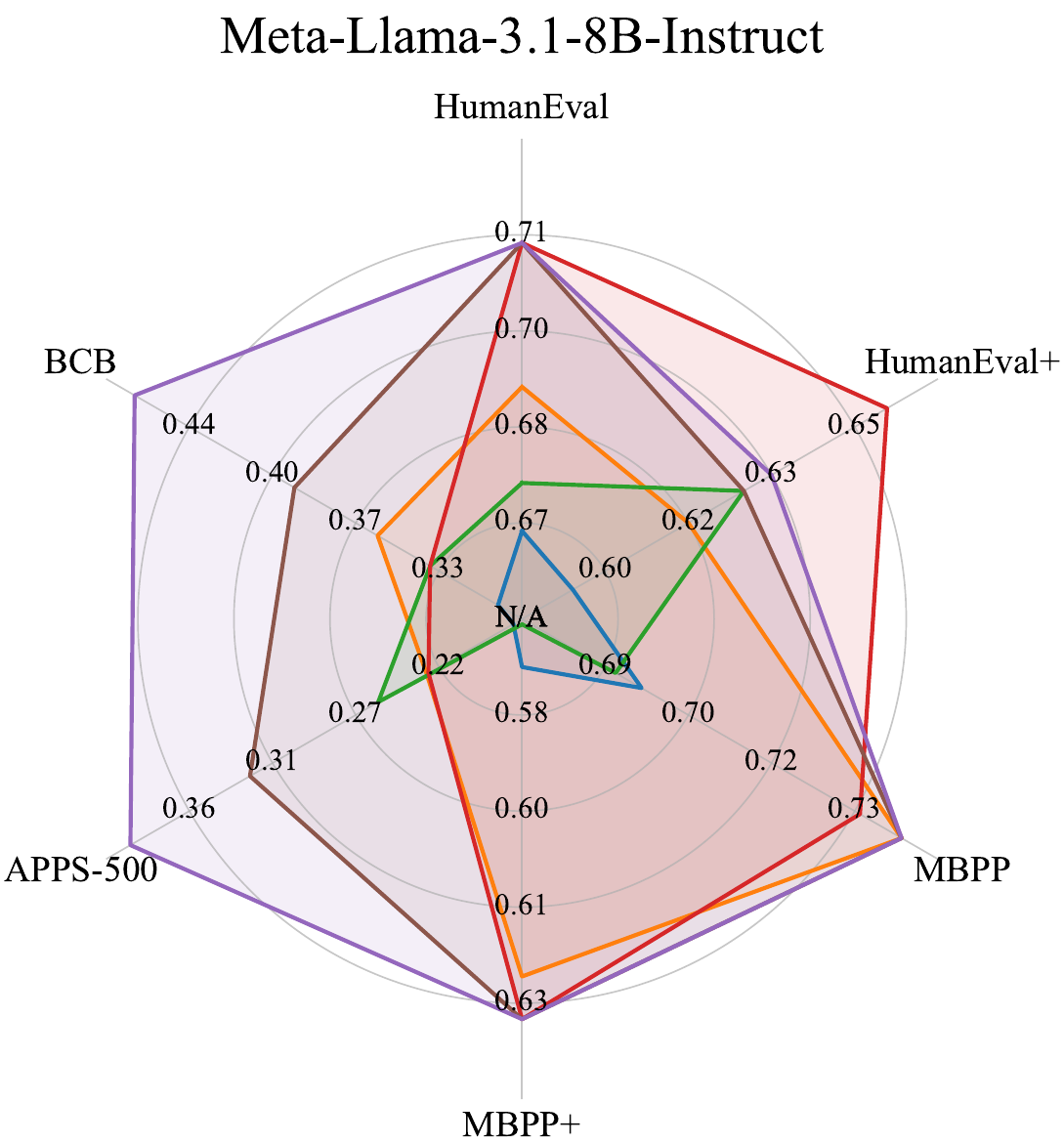}
    \end{subfigure}
    \hfill
    \begin{subfigure}[b]{0.485\textwidth}
        \centering
        \label{fig: DS-Coder}
        \includegraphics[height=1.9in]{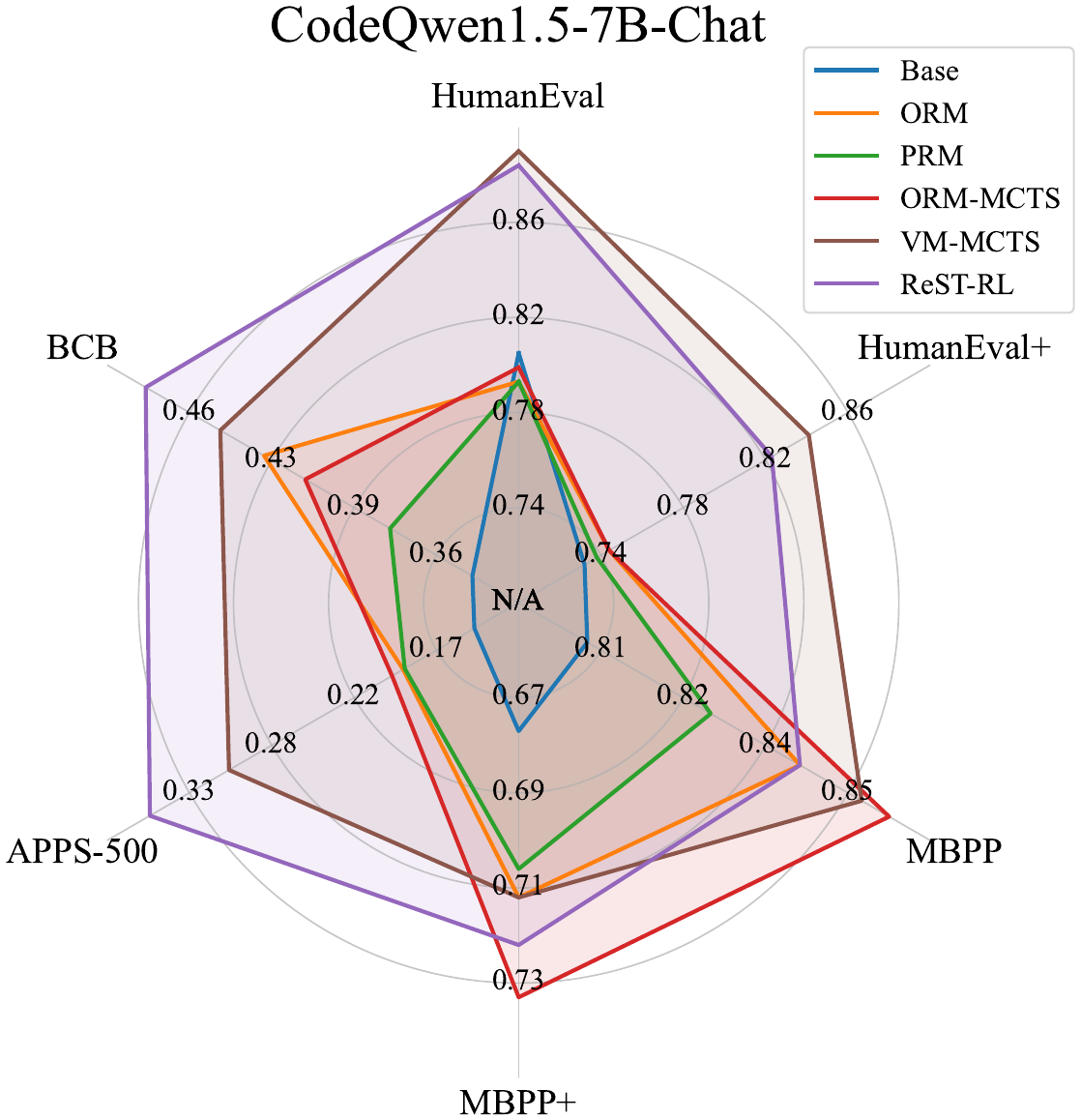}
    \end{subfigure}
    \caption{Performance of different base LLM policies when using \model and verification methods on all benchmarks. All verification is based on $100$ samples, with the sampling temperature set to $0.7$.}
    \label{fig: reward_radar}
\end{figure}

\vpara{Additional results for decoding and verification.} 
Figure~\ref{fig: reward_radar} provides per-benchmark results for the compared decoding and verification methods, and Table~\ref{tab: reward_model_other} reports averages for Llama-3.1-8B-Instruct and CodeQwen1.5-7B-Chat. To control domain mismatch, we also train a Qwen3-8B ORM on $Q_\text{train}$ with the same unit-test reward as the VM. Table~\ref{tab: reward_qwen} shows that this ORM obtains 54.9\% with BoN and 55.0\% with MCTS, versus 68.9\% for the complete \model pipeline.

\begin{table}[t!]
    \centering
    \caption{Average results of \model and different verification methods on all benchmarks for Llama-3.1-8B-Instruct and CodeQwen1.5-7B-Chat.}
    \resizebox{0.5\textwidth}{!}{
    \begin{tabular}{c|c|c}
    \specialrule{.16em}{0pt}{.65ex}
    Method & {Llama-3.1-8B-Instruct} & {CodeQwen1.5-7B-Chat} \\
    \specialrule{.05em}{.4ex}{.65ex}
    Base & 0.435 & 0.502 \\
    \specialrule{.05em}{.4ex}{.65ex}
    ORM & 0.480 & 0.542 \\
    \specialrule{.05em}{.4ex}{.65ex}
    PRM & 0.466 & 0.526 \\
    \specialrule{.05em}{.4ex}{.65ex}
    ORM-MCTS & 0.481 & 0.545 \\
    \specialrule{.05em}{.4ex}{.65ex}
    VM-MCTS & 0.519 & 0.599 \\
    \specialrule{.05em}{.4ex}{.65ex}
    \textbf{ReST-RL} & \textbf{0.556} & \textbf{0.616} \\
    \specialrule{.16em}{.4ex}{0pt}
    \end{tabular}
    }
    \label{tab: reward_model_other}
\end{table}

\begin{table}[t!]
  \centering
  \caption{Detailed inference-time verification results for Qwen3-8B. To ensure domain parity, we include two extra baselines that adopt an ORM trained on the same dataset $Q_{train}$.}
  \resizebox{\textwidth}{!}{
    \begin{tabular}{cl|cccccc|c}
    \specialrule{.16em}{0pt} {.65ex}
    Model & Method & HumanEval & HumanEval+ & MBPP  & MBPP+ & APPS-500 & BCB & Avg. \\
    \specialrule{.05em}{.4ex}{.65ex}
    \multicolumn{1}{c}{\multirow{6}[8]{*}{Qwen3-8B}} 
    & Base & 0.829 & 0.780 & 0.717 & 0.622 & 0.118 & 0.418 & 0.503 \\
    & ORM (Skywork) & 0.841 & 0.787 & 0.746 & 0.661 & 0.165 & 0.441 & 0.531 \\
    & PRM (Qwen) & 0.860 & 0.817 & 0.714 & 0.614 & 0.138 & 0.423 & 0.516 \\
    & ORM (Skywork) + MCTS & 0.854 & 0.805 & 0.746 & 0.661 & 0.193 & 0.424 & 0.538 \\
    & ORM (Trained) & 0.854 & 0.823 & 0.746 & 0.643 & 0.214 & 0.449 & 0.549 \\
    & ORM (Trained) + MCTS & 0.860 & 0.826 & 0.743 & 0.648 & 0.221 & 0.441 & 0.550 \\
    & \modelTest & 0.866 & 0.829 & 0.775 & 0.675 & 0.415 & 0.471 & 0.615 \\
    & \textbf{\model} & 0.866 & 0.829 & 0.817 & 0.704 & 0.642 & 0.506 & \textbf{0.689} \\
    \specialrule{.16em}{.4ex}{0pt}
    \end{tabular}
  }
  \label{tab: reward_qwen}
\end{table}

\subsection{Discussions}
\label{sec: discussion}

\vpara{Practical feasibility of the framework.}
Although \model improves reasoning performance through a two-stage pipeline, its practical feasibility should be assessed from an end-to-end perspective. For the self-training stage, Table~\ref{tab: e2e_train} shows that, even when the pre-sampling cost is included, \modelTrain remains more efficient than naive GRPO in terms of wall-clock time and total GPU hours in the long run. This is because the main bottleneck of RL training lies in repeated online sampling and policy optimization, whereas the offline pre-sampling step is highly parallelizable and incurs only limited additional latency. More importantly, the filtered and partially reassembled training data allow the policy to reach the same or higher performance levels in fewer effective optimization steps, which explains why the end-to-end cost is still reduced despite the extra pre-sampling stage.
For inference, Table~\ref{tab: e2e_test} gives the equal-sample view on the base policy: VM-MCTS is 1.8--2.6 times slower than ORM-BoN, depending on tree shape, while yielding higher APPS-500 accuracy. This comparison uses different verifiers and therefore measures the practical pipeline rather than isolating tree allocation. The causal control in Table~\ref{tab: controlled_results}(c) instead holds the ReST-GRPO policy and VM fixed and approximately matches mean wall-clock: VM-MCTS obtains 0.642 versus 0.619 for VM-BoN at about 41.5 seconds per prompt. VM-MCTS is thus most appropriate when quality warrants additional latency, rather than strict real-time settings.

\begin{table}[t!]
  \centering
  \caption{Equal-sample test-time assessment on APPS-500 using the base Qwen3-8B policy. Each approach generates 100 candidates. BoN uses the coding ORM reported in Table~\ref{tab: reward_qwen}, whereas VM-MCTS uses its corresponding VM; this table reports the practical sample-matched trade-off rather than isolating tree allocation.}
  \resizebox{\linewidth}{!}{
    \begin{tabular}{l|cccc}
    \specialrule{.16em}{0pt}{.65ex}
    Method & ORM-BoN ($N=100$) & VM-MCTS ($n=5, T=20$) & VM-MCTS ($n=10, T=10$) & VM-MCTS ($n=20, T=5$) \\
    \specialrule{.05em}{.4ex}{.65ex}
    Avg. Time (s) / Prompt & 16.0 & 41.5 & 36.4 & 28.6 \\
    APPS-500 & 0.214 & 0.415 & 0.355 & 0.327 \\
    \specialrule{.16em}{.4ex}{0pt}
    \end{tabular}
  }
  \label{tab: e2e_test}
\end{table}

\vpara{Output diversity after training.}
A natural concern for methods that bias training toward high-reward trajectories is whether they may reduce exploration and lead to diversity collapse. The diversity results in Table~\ref{tab: diversity} do not support this concern. ReST-GRPO attains a Self-BLEU score close to the base model and lower than GRPO, indicating that the proposed training procedure does not simply force the policy into a narrow mode. This observation is consistent with the design of \modelTrain: the method does not deterministically replay a single trajectory, but instead reshapes the training distribution by combining prompt-level filtering with partial-state sampling. As a result, optimization is biased toward promising regions of the search space without eliminating variability in subsequent rollouts.
If the gains arose mainly from collapse onto a narrow set of patterns, Self-BLEU would increase sharply. Instead, diversity is preserved while the controlled coding results improve. The preliminary out-of-domain results in Table~\ref{tab: ood} provide additional transfer evidence.

\begin{table}[t!]
  \centering
  \caption{Average Self-BLEU score on APPS-500 for models trained with different approaches. For each prompt, we sample 8 answers to calculate the score. Note that a higher Self-BLEU indicates lower output diversity.}
  \resizebox{0.65\linewidth}{!}{
    \begin{tabular}{l|cccc}
    \specialrule{.16em}{0pt}{.65ex}
    Method & Base & GRPO & ReST-DPO & ReST-GRPO \\
    \specialrule{.05em}{.4ex}{.65ex}
    Avg. Self-BLEU (APPS) & 0.610 & 0.620 & 0.602 & 0.613 \\
    \specialrule{.16em}{.4ex}{0pt}
    \end{tabular}
  }
  \label{tab: diversity}
\end{table}

\vpara{Action granularity.}
In this work, we define an action as a single line of text. We view this choice as a practical compromise between expressiveness, generality, and search efficiency. Compared with token-level actions, line-level actions substantially reduce the branching complexity of tree search and make value estimation more stable, since each transition corresponds to a semantically more meaningful unit. This consideration is also consistent with recent long-horizon code reasoning studies such as LSR-MCTS~\citep{lu2026tokenlineenhancingcode}, which show that salient attention spikes often emerge around line boundaries and suggest that line-level units better capture long-range dependencies in code generation. Compared with coarser alternatives such as full reasoning steps or code blocks, line-level actions are easier to apply across domains because they do not require domain-specific segmentation rules or additional annotations. Recent inference-scaling work also supports this view: DISC~\citep{light2025discdynamicdecompositionimproves} argues that newline-delimited units often correspond to semantically coherent reasoning segments, while ReSCALE~\citep{ugadiarov2026revisitingtreesearchllms} shows that using delimiter-terminated spans instead of token-level actions yields better compute scaling for search-based reasoning.
The same line-level formulation can be applied without domain-specific segmentation rules, and our math/science results provide preliminary evidence that it remains useful beyond code. It is also aligned with process-supervision practice: PRM800K-style steps are commonly organized line by line~\citep{lightman2023letsverifystepstep}. We do not claim that this granularity is universally optimal; adaptive or task-specific action abstractions remain an important direction.

\vpara{Comparison with other tree search methods.}
As summarized in Table~\ref{tab: mcts_comparison}, \modelTest differs from related tree-search methods in how its guidance signal is trained and used. PPO-MCTS~\citep{liu2024dontthrowawayvalue} learns policy and value jointly, whereas we train a dedicated static-policy VM after policy optimization. Math-Shepherd~\citep{wang2024mathshepherdverifyreinforcellms} and ReST-MCTS$^*$~\citep{zhang2024restmctsllmselftrainingprocess} focus on process-oriented supervision and search-based data generation, whereas our target is expected terminal reward for a partial state. LATS~\citep{zhou2024languageagenttreesearch} instead emphasizes interaction, reflection, and external feedback.
Moreover, the motivation for introducing \modelTest is not merely to add another MCTS variant, but to complement the weakness of GRPO-style policy optimization. GRPO is efficient, yet it removes explicit reward feedback learning from the RL loop, which may weaken fine-grained guidance during both training and inference. Our framework reintroduces that missing signal in a lightweight manner. \modelTrain improves the policy-induced trajectory distribution, and \modelTest exploits this improved distribution to learn value estimates that are critical at inference. In this sense, the comparison with existing MCTS-based methods also clarifies the conceptual contribution of \model. It is an attempt to reconnect policy optimization and value-guided reasoning without returning to the full cost of online actor-critic learning.

\begin{table}[t]
\centering
\caption{Design comparison with representative MCTS-based approaches.}
\label{tab: mcts_comparison}
\setlength{\tabcolsep}{3.5pt}
\renewcommand{\arraystretch}{1.18}
\begin{tabular}{p{2.0cm}p{2.75cm}p{2.45cm}p{2.45cm}p{2.7cm}}
\specialrule{.16em}{0pt}{.65ex}
\textbf{Feature} & \textbf{ReST-RL (VM-MCTS)} & \textbf{PPO-MCTS} & \textbf{ReST-MCTS$^*$} & \textbf{LATS} \\
\specialrule{.05em}{.4ex}{.65ex}
Guidance signal & State-value model trained on self-collected MCTS data & Token-level value network from PPO & Process reward model trained with search-derived targets & Environment feedback and self-reflection \\
\specialrule{.03em}{.3ex}{.5ex}
Training coupling & Static-policy VM training after policy optimization & Joint policy--value optimization & Iterative self-training with PRM-guided search & Agent search without policy--value training \\
\specialrule{.03em}{.3ex}{.5ex}
Action granularity & Line-level & Typically token-level & Reasoning-step level & Task/agent-action level \\
\specialrule{.03em}{.3ex}{.5ex}
Primary use & Value-guided allocation and final selection & Value-guided text decoding & Search-based data generation & Interactive reasoning and planning \\
\specialrule{.16em}{.4ex}{0pt}
\end{tabular}
\end{table}

\subsection{Related Work}
\label{sec: related_work}
\vpara{LLMs for reasoning and code synthesis.}
Recent advancements have facilitated the growing application of LLMs to reasoning tasks like coding \citep{Li_2022, guo2024deepseekcoderlargelanguagemodel,codeqwen1.5,zheng2024opencodeinterpreter} and math \citep{openai2024gpt4technicalreport, shao2024deepseekmathpushinglimitsmathematical}. To assist LLMs on code synthesis, previous approaches use test cases to provide reward signals \citep{zhang2023planninglargelanguagemodels}. However, though demonstrating notable performance on elementary programming tasks, LLMs continue to struggle with complex ones, such as those found in competitive programming platforms like Codeforces \citep{hendrycks2021measuringcodingchallengecompetence}.

\vpara{Selective self-training, curriculum, and policy optimization.}
Self-training methods such as STaR, ReST, and ReST-EM sample from the current policy and train on filtered complete responses \citep{zelikman2022starbootstrappingreasoningreasoning,gulcehre2023reinforcedselftrainingrestlanguage,singh2023beyond}. Difficulty-aware methods instead control which original problems are presented: E2H Reasoner uses an easy-to-hard cosine curriculum \citep{parashar2025curriculum}, while DAPO dynamically resamples groups with nonzero reward variance \citep{yu2025dapoopensourcellmreinforcement}. ReST-GRPO shares their goal of improving weak training signals but intervenes at a different level. Selected partial states become new online-GRPO contexts and their suffixes are re-sampled, rather than using complete accepted responses as imitation targets or only rescheduling original prompts. Table~\ref{tab: component_ablation} directly compares these levels of intervention.

\vpara{Reasoning-time search.}
Prompting methods such as Chain-of-Thought \citep{wei2023chainofthoughtpromptingelicitsreasoning} and tree search methods such as Tree-of-Thought \citep{yao2023treethoughtsdeliberateproblem} improve decoding without matched policy updates. MCTS-based methods further allocate computation using rewards, learned values, or reflection \citep{liu2024dontthrowawayvalue,zhang2024restmctsllmselftrainingprocess,xu2025sramctsselfdrivenreasoningaugmentation,zhang2024accessinggpt4levelmathematical}. Our controlled VM-BoN comparison holds both policy and verifier fixed, isolating the benefit of using the learned VM during allocation from its use only at final selection.

\vpara{Output verification with reward models.}
Outcome reward models score complete outputs, whereas process reward models provide intermediate feedback \citep{uesato2022solvingmathwordproblems,lightman2023letsverifystepstep}. PRMs can improve verification but require high-quality process targets. Monte-Carlo supervision reduces annotation cost \citep{wang2024mathshepherdverifyreinforcellms,zhang2024restmctsllmselftrainingprocess}, although the resulting targets can be noisy \citep{zhang2025lessonsdevelopingprocessreward}. We use the term Value Model because our target is explicitly $V^\pi(s)$---expected terminal reward from a partial state under a static policy---rather than step correctness.


\subsection{Limitations}
\label{sec: limitation}
First, \model does not study repeated multi-round coupling of an updated policy and VM beyond the reported two-stage pipeline. Second, VM-MCTS has higher latency than BoN at equal candidate counts. The approximately latency-matched control is reported on Qwen3-8B and APPS-500 and does not cover every deployment regime. Third, the controlled component and matched-VM runs use one training seed and one base model, although the main ReST-GRPO result is supported by four independent runs and multiple model families. Finally, the math/science experiments use Qwen3-8B and show transfer without target-domain tuning, but they do not establish a universal reasoning value model.

\subsection{Table of Hyperparameters}
\label{sec: hyperparameters}

\begin{table}[ht!]
    \centering
    \caption{Main hyperparameters and default values for ReST-GRPO.}
    \label{tab: rest_grpo_params}
    \resizebox{\textwidth}{!}{
    \begin{tabular}{c|c|c|c|c|c}
    \specialrule{.16em}{0pt}{.65ex}
    Hyperparameter & $N$ & $\sigma_0$ & $r_0$ & $\alpha$ & $\beta$ \\
    \specialrule{.05em}{.4ex}{.65ex}
    Meaning & Number of solution samples & SD threshold & Reward threshold & Sampling exponent factor & Sample ratio \\
    \specialrule{.05em}{.4ex}{.65ex}
    Default Value & 30 & 0.05 & 0.9 & 0.95 & 0.5 \\
    \specialrule{.16em}{.4ex}{0pt}
    \end{tabular}
    }
\end{table}

\begin{table}[ht!]
    \centering
    \caption{Main hyperparameters and default values for VM-MCTS.}
    \label{tab: vm_mcts_params}
    \resizebox{\textwidth}{!}{
    \begin{tabular}{c|c|c|c|c}
    \specialrule{.16em}{0pt}{.65ex}
    Hyperparameter & $T$ & $n$ & $c$ & $\epsilon$ \\
    \specialrule{.05em}{.4ex}{.65ex}
    Meaning & Search iterations & Number of branches expanded & Exploration constant & Constant avoiding zero-division in UCT \\
    \specialrule{.05em}{.4ex}{.65ex}
    Default Value & 30 & 5 & 0.4 & 0.1 \\
    \specialrule{.16em}{.4ex}{0pt}
    \end{tabular}
    }
\end{table}

\clearpage
\subsection{Mathematical Proof for Value Estimation}
\label{sec: proof}
In this part we prove that our proposed value-based MC rollout tends to have more precise estimation of partial state's value, in terms of variance. Suppose we are estimating the value of partial state $S_i=(q,a_{1,2,\dots,i})$ for a fixed policy $\pi$. We are allowed to use two verifier models, a value model $V_\phi$ and a reward model $R_\lambda$ (which is allowed to be identical to $V_\phi$), while the actual reward function $R$ and value function $V$ are unknown (since it's test time). As a reasonable approximation, we assume that the two models can be regarded as two unbiased estimator variables that have the same variance when predicting, along with some conditions of independence:

\begin{equation}
\label{eq: estimator_approximation}
    \begin{aligned}
    &V_\phi(S)=V(S)+\epsilon_V, \\
    &R_\lambda(S_{end})=R(S_{end})+\epsilon_R, \\
    s.t.\ &\mathbb{D}[\epsilon_V]=\mathbb{D}[\epsilon_R],\mathbb{E}[\epsilon_V]=\mathbb{E}[\epsilon_R]=0, \\
    &\epsilon_V\text{ is independent of }S, \\
    &\epsilon_R\text{ is independent of }S_{end}.
    \end{aligned}
\end{equation}
Now, we consider two approaches for value estimation, with a limit of $n$ simulations. For value-based MC rollout, it does simulation with the random variable $S_{i+1}$, i.e. the state reached by $\pi$ performing a single action. It estimates $V(S_i)$ by:
\begin{equation}
\begin{aligned}
    &\hat{V}_{v}(S_i)=\frac{1}{n}\sum_{j=1}^n V_\phi(S_{i+1}^{(j)}), \\
    s.t.\ &S_{i+1}^{(1)},S_{i+1}^{(2)},\dots,S_{i+1}^{(n)}\overset{\text{i.i.d.}}{\sim}\pi(S_{i+1}|S_i).
\end{aligned}
\end{equation}
On the other hand, regular complete rollout does simulation with the random variable $S_{end}$, i.e. the end state reached after a complete trace is generated. It estimates $V(S_i)$ by:
\begin{equation}
\begin{aligned}
    &\hat{V}_{r}(S_i)=\frac{1}{n}\sum_{j=1}^n R_\lambda(S_{end}^{(j)}), \\
    s.t.\ &S_{end}^{(1)},S_{end}^{(2)},\dots,S_{end}^{(n)}\overset{\text{i.i.d.}}{\sim}\pi(S_{end}|S_i).
\end{aligned}
\end{equation}
We now state the proposition to be proved:
\begin{equation}
\label{eq: expectation_claim}
    \mathbb{E}[\hat{V}_{r}(S_i)]=\mathbb{E}[\hat{V}_{v}(S_i)]=V(S_i),
\end{equation}
and
\begin{equation}
\label{eq: variance_claim}
    \mathbb{D}[\hat{V}_{r}(S_i)]\geq\mathbb{D}[\hat{V}_{v}(S_i)].
\end{equation}
If the statements are true, we can conclude that $\hat{V}_{v}(S_i)$ is a better estimator than $\hat{V}_{r}(S_i)$ due to its smaller variance, which justifies our original claim. Now, we provide the proof for Eq.~(\ref{eq: expectation_claim}).
\begin{proof}
    First, we notice that by Eq.~(\ref{eq: estimator_approximation}):
    \[
    \begin{aligned}
    \hat{V}_{r}(S_i)&=\frac{1}{n}\sum_{j=1}^n R_\lambda(S_{end}^{(j)}) \\
    &=\frac{1}{n}\sum_{j=1}^n  (R(S_{end}^{(j)})+\epsilon_R) \\
    &=\epsilon_R+ \frac{1}{n}\sum_{j=1}^n  R(S_{end}^{(j)})
    \end{aligned}
    \]
    Substituting the above equation into the expression of expectation, we derive that:
    \[
    \begin{aligned}
    \mathbb{E}[\hat{V}_{r}(S_i)]&=\mathbb{E}[\epsilon_R+ \frac{1}{n}\sum_{j=1}^n  R(S_{end}^{(j)})] \\
    &=\mathbb{E}[\epsilon_R]+\frac{1}{n}\sum_{j=1}^n  \mathbb{E}[R(S_{end}^{(j)})] \\
    &=\frac{1}{n}\sum_{j=1}^n  \mathbb{E}[R(S_{end})|S_i] \\
    &=\mathbb{E}[R(S_{end})|S_i] \\
    &=V(S_i)
    \end{aligned}
    \]
    Similarly, we can know that:
    \[
    \begin{aligned}
    \hat{V}_{v}(S_i)&=\frac{1}{n}\sum_{j=1}^n V_\phi(S_{i+1}^{(j)}) \\
    &=\frac{1}{n}\sum_{j=1}^n  (V(S_{i+1}^{(j)})+\epsilon_V) \\
    &=\epsilon_V+ \frac{1}{n}\sum_{j=1}^n  V(S_{i+1}^{(j)})
    \end{aligned}
    \]
    Thus, we can finally derive the desired properties:
    \[
    \begin{aligned}
    \mathbb{E}[\hat{V}_{v}(S_i)]&=\mathbb{E}[\epsilon_V+ \frac{1}{n}\sum_{j=1}^n  V(S_{i+1}^{(j)})] \\
    &=\mathbb{E}[\epsilon_V]+\frac{1}{n}\sum_{j=1}^n  \mathbb{E}[V(S_{i+1}^{(j)})] \\
    &=\frac{1}{n}\sum_{j=1}^n  \mathbb{E}[V(S_{i+1})|S_i] \\
    &=\mathbb{E}[V(S_{i+1})|S_i] \\
    &=V(S_i)
    \end{aligned}
    \]
\end{proof}
Next, we provide proof for Eq.~(\ref{eq: variance_claim}) as follows.
\begin{proof}
    Using the conclusions made by the proof above, we can derive the required property:
    \[
    \begin{aligned}
    &\mathbb{D}[\hat{V}_{r}(S_i)]\geq\mathbb{D}[\hat{V}_{v}(S_i)] \\
    \impliedby &\mathbb{D}[\epsilon_R+ \frac{1}{n}\sum_{j=1}^n  R(S_{end}^{(j)})]\geq \mathbb{D}[\epsilon_V+ \frac{1}{n}\sum_{j=1}^n  V(S_{i+1}^{(j)})] \\
    \impliedby &\mathbb{D}[\epsilon_R]+ \mathbb{D}[\frac{1}{n}\sum_{j=1}^n  R(S_{end}^{(j)})]\geq \mathbb{D}[\epsilon_V]+\mathbb{D}[\frac{1}{n}\sum_{j=1}^n  V(S_{i+1}^{(j)})] \\
    \impliedby &\frac{1}{n^2}\mathbb{D}[\sum_{j=1}^n  R(S_{end}^{(j)})]\geq \frac{1}{n^2}\mathbb{D}[\sum_{j=1}^n  V(S_{i+1}^{(j)})] \\
    \impliedby &\sum_{j=1}^n\mathbb{D}[R(S_{end}^{(j)})]\geq \sum_{j=1}^n \mathbb{D}[V(S_{i+1}^{(j)})] \\
    \impliedby &\mathbb{D}[R(S_{end})|S_i]\geq \mathbb{D}[V(S_{i+1})|S_i] \\
    \impliedby &\mathbb{E}[(R(S_{end}))^2|S_i]-(\mathbb{E}[R(S_{end})|S_i])^2\geq \mathbb{E}[(V(S_{i+1}))^2|S_i]-(\mathbb{E}[V(S_{i+1})|S_i])^2 \\
    \impliedby &\mathbb{E}[(R(S_{end}))^2|S_i]\geq \mathbb{E}_{S_{i+1}\sim \pi(S_{i+1}|S_i)}[(V(S_{i+1}))^2|S_i] \\
    \impliedby &\mathbb{E}_{S_{i+1}\sim \pi(S_{i+1}|S_i)}[\mathbb{E}[(R(S_{end}))^2|S_{i+1}]|S_i]\geq \mathbb{E}_{S_{i+1}\sim \pi(S_{i+1}|S_i)}[(\mathbb{E}[R(S_{end})|S_{i+1}])^2|S_i] \\
    \impliedby &\mathbb{E}_{S_{i+1}\sim \pi(S_{i+1}|S_i)}[\mathbb{E}[(R(S_{end}))^2|S_{i+1}]]\geq \mathbb{E}_{S_{i+1}\sim \pi(S_{i+1}|S_i)}[(\mathbb{E}[R(S_{end})|S_{i+1}])^2] \\
    \impliedby &\mathbb{E}_{S_{i+1}\sim \pi(S_{i+1}|S_i)}[\mathbb{E}[(R(S_{end}))^2|S_{i+1}]-(\mathbb{E}[R(S_{end})|S_{i+1}])^2]\geq 0 \\
    \impliedby &\mathbb{E}_{S_{i+1}\sim \pi(S_{i+1}|S_i)}[\mathbb{D}[R(S_{end})|S_{i+1}]]\geq 0 \\
    \impliedby &\mathbb{D}[R(S_{end})|S_{i+1}]\geq 0
    \end{aligned}
    \]
\end{proof}





\end{document}